\documentclass[journal]{IEEEtran}
\usepackage{amsmath,amsfonts}
\usepackage{algorithmic}
\usepackage{algorithm}
\usepackage{array}
\usepackage[caption=false,font=normalsize,labelfont=sf,textfont=sf]{subfig}
\usepackage{textcomp}
\usepackage{stfloats}
\usepackage{url}
\usepackage{verbatim}
\usepackage{graphicx}
\usepackage{cite}
\usepackage{hyperref}
\hypersetup{hidelinks,
	colorlinks=true,
	pdfstartview=Fit,
	breaklinks=true}
\usepackage{multirow}
\usepackage{arydshln}
\usepackage{threeparttable}
\begin{document}

\title{Shared Coupling-bridge for Weakly Supervised Local Feature Learning}

\author{Jiayuan Sun, Jiewen Zhu, Luping Ji$^{\ast}$
\thanks{*Corresponding author.}
\thanks{This work is supported by the National Natural Science Foundation of China (NSFC) under Grant No.
61972072.}
\thanks{Jiayuan Sun, Jiewen Zhu, Luping Ji are with the School of Computer
Science and Engineering, University of Electronic Science and Technology
of China, Chengdu 611731, China}
}

\maketitle

\begin{abstract}
Sparse local feature extraction is usually believed to be of important significance in typical vision tasks such as simultaneous localization and mapping, image matching and 3D reconstruction. 
At present, it still has some deficiencies needing further improvement, mainly including the discrimination power of extracted local descriptors, the localization accuracy of detected keypoints, and the efficiency of local feature learning.
This paper focuses on promoting the currently popular sparse local feature learning with camera pose supervision. 
Therefore, it pertinently proposes a Shared Coupling-bridge scheme with four light-weight yet effective improvements for weakly-supervised local feature (SCFeat) learning. 
It mainly contains: 
i) the \emph{Feature-Fusion-ResUNet Backbone} (F2R-Backbone) for local descriptors learning, 
ii) a shared coupling-bridge normalization to improve the decoupling training of description network and detection network,
iii) an improved detection network with peakiness measurement to detect keypoints and 
iv) the fundamental matrix error as a reward factor to further optimize feature detection training. 
Extensive experiments prove that our SCFeat improvement is effective. 
It could often obtain a state-of-the-art performance on classic image matching and visual localization. In terms of 3D reconstruction, it could still achieve competitive results. For sharing and communication, our source codes are available at \href{https://github.com/sunjiayuanro/SCFeat.git}{https://github.com/sunjiayuanro/SCFeat.git}.
\end{abstract}


\section{Introduction}


\IEEEPARstart{O}{ne} of the necessary steps in many vision tasks is to find pixel correspondence. Sparse local feature, as one primary clue to determine correspondence, has been widely explored in many areas, such as simultaneous localization and mapping (SLAM) \cite{9261135,9531062,9173732}, image matching \cite{9527141,9447920} and 3D reconstruction \cite{9817616,6359953,8962030}. It could offer a series of advantages to fulfil pixel correspondence. For example, it can be matched efficiently via the nearest neighbor search strategy and classic Euclidean distance. It also offers a memory-efficient representation, and thus enable some schemes such as visual localization to be smoothly implemented \cite{D2net}.

The classical approaches for sparse local features are usually based on a two-stage pipeline: detecting interest points (\emph{Keypoints}) and then computing the local descriptors for each point. Early works \cite{hcrt:21,BRISK} focus on detection. And they are proposed to distinguish distinctive areas for detecting effective keypoints. Most of them often work well in practice, $e.g.$, the corners-based method \cite{Harris1988ACC} and blobs-based ones \cite{hcrt:21,hcrt:28}. Later methods pay their attentions to the description step and try to design powerful descriptors \cite{SURF,ORB,BRIEF}. Despite obtaining obvious success, these handcrafted feature methods are often limited by the prior knowledge that researchers have accumulated for a special task. For these kinds of methods, therefore it's very crucial to automatically determine the most suitable feature extraction process and feature representation to given images \cite{R2D2}.

\begin{figure}[t]
\centering
\includegraphics[width=0.99\columnwidth]{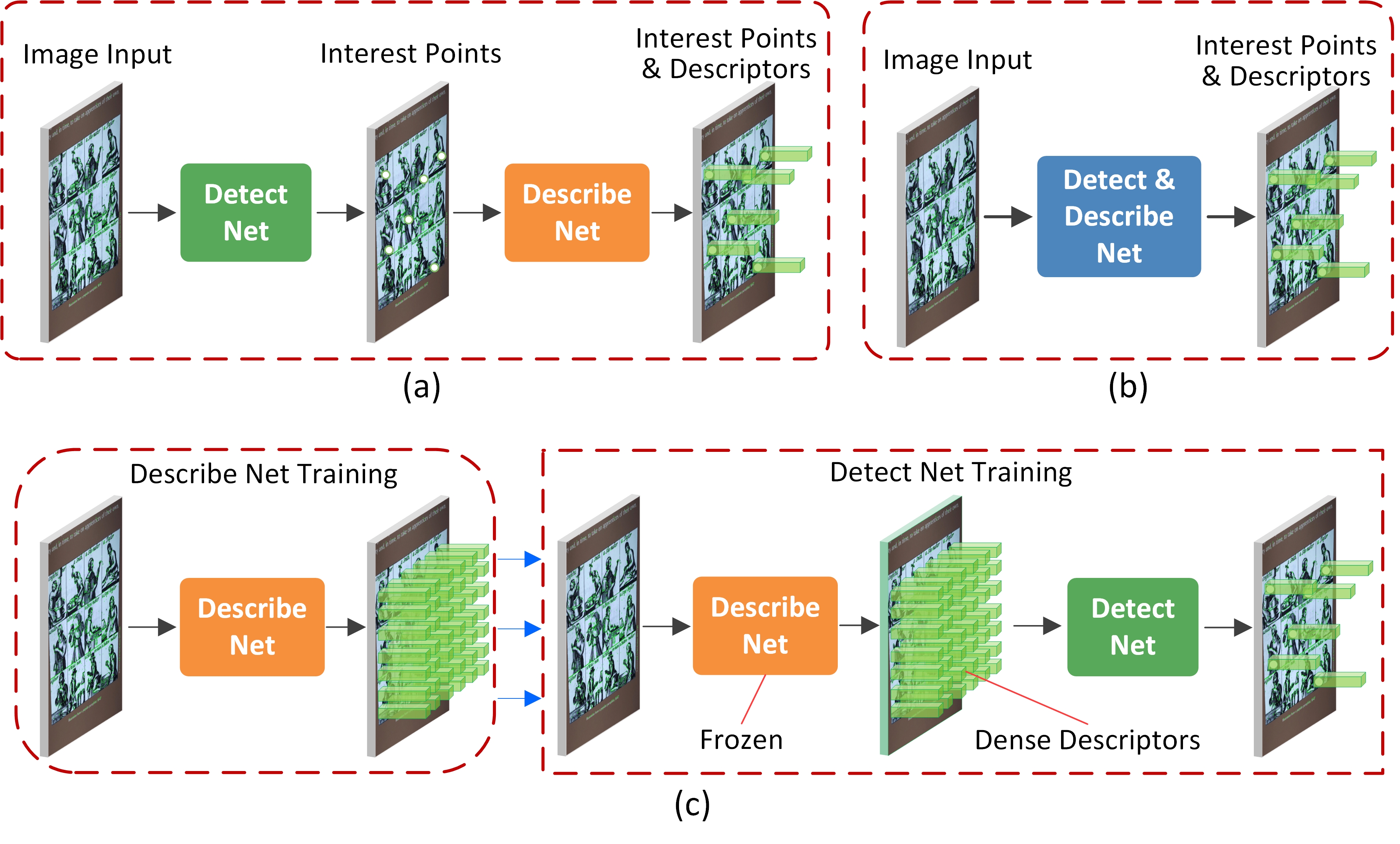}
\caption{The pipeline comparison of different popular pipelines for learnable local feature extraction: (a) two-stage one that first detects interest points and then computes a local descriptor for each point, (b) a single CNN that extracts dense features that serve as both descriptors and detectors, and  (c) a describe-then-detect pipeline with decoupled training, a description network being frozen to produce the dense descriptor for detection training after description learning.}
\label{fig1}
\end{figure}

In the past years, convolutional neural networks have been proved to be superior to hand-engineered representations on almost all images processing tasks \cite{SuperPoint}. Motivated by the success of deep learning schemes, \cite{cndesc,LIFT,LF-Net,SOSNet,ContextDesc,caps} regard sparse local feature extraction as two separate problems, and then replace the detection or description of detect-then-describe pipeline with deep CNNs. Fig. \ref{fig1} (a) illustrates the common structure of this pipeline. Recent works \cite{D2net,R2D2,aslfeat} find that keypoints and descriptors are usually interdependent, so propose a joint training detect-and-describe pipeline. Fig. \ref{fig1} (a) and \ref{fig1} (b) illustrate a difference of this pipeline with a detect-then-describe one. By jointly optimizing description network and detection network, the training scheme on detect-and-describe architecture could achieve better feature extraction performance than a detect-then-describe pipeline. However, most detect-and-describe methods are fully supervised, strongly relying on the dense labels of ground-truth correspondence in training. However, it is usually very difficult to collect the sufficient training samples with dense labels.

\begin{figure}[t]
\centering
\includegraphics[width=0.99\columnwidth]{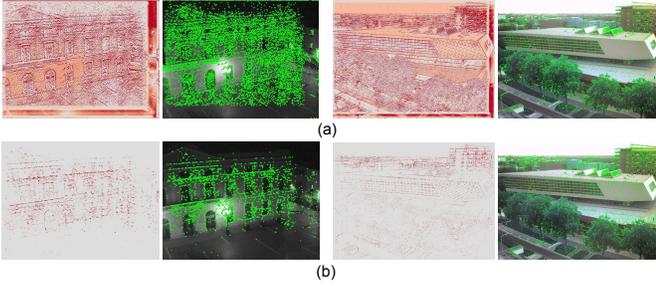} 
\caption{An illustration of the influence for F2R-Backbone. (a) With standard ResUNet Backbone, PoSFeat produces many streaks in dense feature maps and then affects the score maps (Streaks appear at the bottom-margin region and right-margin region of left sub-images.). (b) With our F2R-Backbone, SCFeat generates better dense feature maps and score maps.}
\label{fig2}
\end{figure}

Owing to the convenience of collecting camera pose images, CAPS \cite{caps} proposes to learn descriptors solely from relative camera poses between image pairs, achieving impressive results through a detect-then-describe pipeline. DISK \cite{disk} integrates a weakly-supervised learning into a joint training describe-then-detect pipeline. However, with purely weakly-supervised one, the losses generated by detection network and description network cannot be often distinguished when these two are optimized by jointly training in a describe-then-detect pipeline. Thus, PoSFeat \cite{li2022decoupling} proposes an opposite describe-then-detect pipeline with a decoupled training by weakly supervised feature learning. As illustrated in Fig. \ref{fig1} (c), two networks are separately trained, therefore the loss functions of them are naturally decoupled to overcome possible confusion. This kind of pipeline could achieve better performance than detect-then-describe one, although incurring interference streaks as shown in Fig. \ref{fig2} (a) frequently.

Although these recent works, mentioned above, have obtained impressive performance promotion, there still exist some deficiencies needing further improvement, mainly including the discrimination power of extracted local descriptors, the localization accuracy of detected keypoints, and the efficiency of local feature learning. Therefore, in this paper, motivated by \cite{li2022decoupling,caps,cndesc,aslfeat}, we propose an augmented improvement scheme with only a camera pose supervision for sparse local feature learning. 

In our new scheme ($i.e.$, SCFeat), an F2R-Backbone is designed for weakly-supervised descriptors learning. In local descriptors learning, we additionally propose a shared coupling-bridge normalization for more efficient description network training. Meanwhile, our shared coupling-bridge is also used for more efficient detection network learning. Furthermore, an augmented detection network with peakiness measurement is devised for more accurate keypoints localization.
Besides, we employ a fundamental matrix error as a new reward factor to enhance feature detection training.

The main contributions in SCFeat could be further summarized as follows:
\begin{itemize}
\item{An F2R-Backbone for local descriptors learning is proposed. It's simple yet efficient in promoting the performance of weakly-supervised local feature learning.}

\item{A shared coupling-bridge cross normalization scheme to improve the decoupling training of description network and detection network for highly-efficient local feature learning is devised.}

\item{An augmented detection network with peakiness measurement to detect keypoints is proposed. It efficiently makes our feature detection network achieve a high detection accuracy.}

\item{It's the first that a fundamental matrix error is utilized as an extra reward factor of loss function in optimizing network training.}

\item{Extensive comparison experiments and ablation analysis are done on three open datasets \cite{HPatches,aachen,ETH}. Our improvement obtains obvious performance promotion.}
\end{itemize}

\section{Related Works}\label{relatedworks}

\subsection{High-level Sparse Local Features}

Sparse local features play a crucial role in many computer vision tasks \cite{1642666,ORBSLAM3_TRO}. As a classic approach for sparse local feature extraction, the detect-then-describe pipeline first performs detection and then extracts a feature descriptor from a patch centered around each keypoint. Early efforts
\cite{Harris1988ACC,hcrt:21,SURF,BRISK,ORB,BRIEF} usually adopt hand-crafted detectors and descriptors. Later works \cite{c:29,Quad-Networks} pay their attention to either descriptor or detector replacing, as well as both detector and descriptor learning \cite{LIFT,LF-Net}. However, their feature detectors often consider only small image regions and usually focus on low-level image structures.


Recent-year works find that keypoints and descriptors are often interdependent. In view of this, SuperPoint \cite{SuperPoint} devises a self-supervised paradigm with a bootstrap training strategy to train a model to detect keypoints and extract descriptors jointly. R2D2 \cite{R2D2} deploys effective loss functions to consider both the repeatability and reliability in interest point detection. D2Net \cite{D2net} and ASLFeat \cite{aslfeat} adopt a joint training detect-and-describe pipeline. They first extract dense descriptors and then detect the keypoints from dense descriptors by special rules. Compared with detect-then-describe pipeline, joint training detect-and-describe pipeline 
uses feature maps from deeper layers of a CNN, enabling us to base both feature detection and description on higher-level information.
Therefore, it belongs to higher-level structures and its descriptors are locally-unique ones. Meanwhile, dense descriptors involve richer image context, so they could often generate better performance \cite{cndesc}.

\subsection{Weakly Supervised Local Feature Learning}\label{relatedworksB}
Most of the methods above such as R2D2 \cite{R2D2}, D2Net \cite{D2net} and ASLFeat \cite{aslfeat} are fully supervised, and they strongly rely on the dense labels of ground-truth correspondence in training. However, collecting a large dataset with pixel-level ground truth correspondences is usually expensive. Therefore, to reduce dataset plague, self-supervised and weakly supervised learning are further investigated for network training. For example, SuperPoint \cite{SuperPoint} uses a virtual homography to generate an image pair from a single image to fulfil self-supervised learning, achieving obvious improvement results. Nevertheless, simple homography transformations may not always work in real cases. For better adaptability \& stability, weakly-supervised learning has currently attracted broad research enthusiasm in local feature learning.

Since camera poses are easy to collect, CAPS \cite{caps} proposes a weakly-supervised scheme to learn descriptors solely from the relative camera poses between image pairs. In this scheme, it designs an epipolar loss for descriptor learning. In recent works, DISK \cite{disk} also integrates a weakly-supervised learning into a joint training describe-then-detect pipeline by adopting policy gradient. However, when DISK is directly trained by a weakly-supervised loss, obvious performance drop is often observed on pixel-wise metrics. 

Furthermore, PoSFeat \cite{li2022decoupling} claims that weakly-supervised loss cannot often distinguish the possible errors generated differently by inaccurate descriptors and false keypoints. This kind of ambiguity will hinder jointly-training detect-and-describe pipelines to learn better local features. Therefore, it proposes a decoupled-training describe-then-detect pipeline for weakly supervised local feature learning. Its new variation of architecture is briefly illustrated in Fig. \ref{fig1} (c). In this pipeline, the detection network, decoupled from description network, is postponed until highly discriminative and robust descriptors have been obtained. 

\subsection{Normalization in Local Descriptors}
Normalization has been widely used almost in all computer vision tasks \cite{BN,GN}. Early normalizing inputs are only regarded as one of the efficient tricks in training neural networks \cite{645754.668382}. The success of normalization methods may be attributed to reducing the internal covariate shift of hidden layers \cite{BN,LayerN}, changing the loss surface and preventing divergence \cite{3305381.3305417}, and accelerating training \cite{BN}. Different from the methods above, L2 normalization is often used to generate a normalized representation vector at the end of description network. HyNet \cite{HyNet} finds that L2 normalization enables local descriptors to obtain better description ability and could lead to a performance improvement in matching accuracy. L2 normalization is essential in recent works \cite{SuperPoint,disk}. Similarly, even hand-crafted descriptors could also often benefit from the use of L2 normalization \cite{HPatches}. 

Moreover, a recent new work CNDesc \cite{cndesc} thinks that L2 normalization also potentially causes a normalized descriptor to lose some specific information ($i.e.$, L2 norm). It has a denser distribution in description space. This kind of distribution may impair the distinguish ability of learned local descriptors. Thus, motivated by learnable Z-score normalization, CNDesc proposes a learnable cross normalization as an alternative to L2 normalization.
It normalizes the feature descriptors in both spatial and channel dimensions with adaptive parameters, to retain distinguishing information. 

\begin{table}[h]
\Large
\centering

\caption{Some Important Notations in this paper.
}
\renewcommand\arraystretch{1.2}
\resizebox{.99\columnwidth}{!}{
\begin{tabular}{ll}
    \hline
    Symbols & Meaning Explanations\\
    \hline
    $\oplus$    & The concatenate operation. \\
    $\otimes$   & The Hadamard product.\\
    $H$ & The height of given image. \\
    $W$ & The width of given image. \\
    $C$ & The number of channels in feature maps. \\
    $D$ & The dense descriptor maps, $D \in \mathbb{R}^{H\times W \times C}$ \\
    $\mathcal{H}(\cdot)$ & The composite function of operations. \\
    $\mathcal{F}_{(\cdot)}$ & The dense feature maps output from $(\cdot)$ layer. \\
    $\boldsymbol{x}_{i},\boldsymbol{y}_{i}$ & The $i$-th query point in the vector of keypoints. \\
    $e(\boldsymbol{x}_{i}),e(\boldsymbol{y}_{i})$ & The epipolar line linking query point $\boldsymbol{x}_{i}$ and $\boldsymbol{y}_{i}$. \\
    $\boldsymbol{\rm y}^{k}$ & The feature map of the $k$-th channel in dense feature maps. \\
    $\eta_{ij}^{k}$ & The channel-wise score at the location $(i,j)$ of a feature map  \\
    & in the $k$-th channel. \\
    $\zeta_{ij}^{k}$ & The local score at the location $(i,j)$ of a feature map in the \\
    & $k$-th channel.\\ 
    $F$ & The Ground-Truth value of fundamental matrix. \\
    $\hat{F}$ & The estimation of fundamental matrix. \\
    $N$ & The kernel size for non-maximum suppression. \\
    $\epsilon_{s}$ & The score threshold for a score map (the keypoints whose \\& scores smaller than $\epsilon_{s}$ will be filtered). \\
    $\epsilon_{r}$ & The threshold of the ratio test. \\
    
    \hline
\end{tabular}
}
\label{table-symbols}
\end{table}

\section{Proposed Methodology}

\begin{figure*}[ht]
\centering
\includegraphics[width=0.99\textwidth]{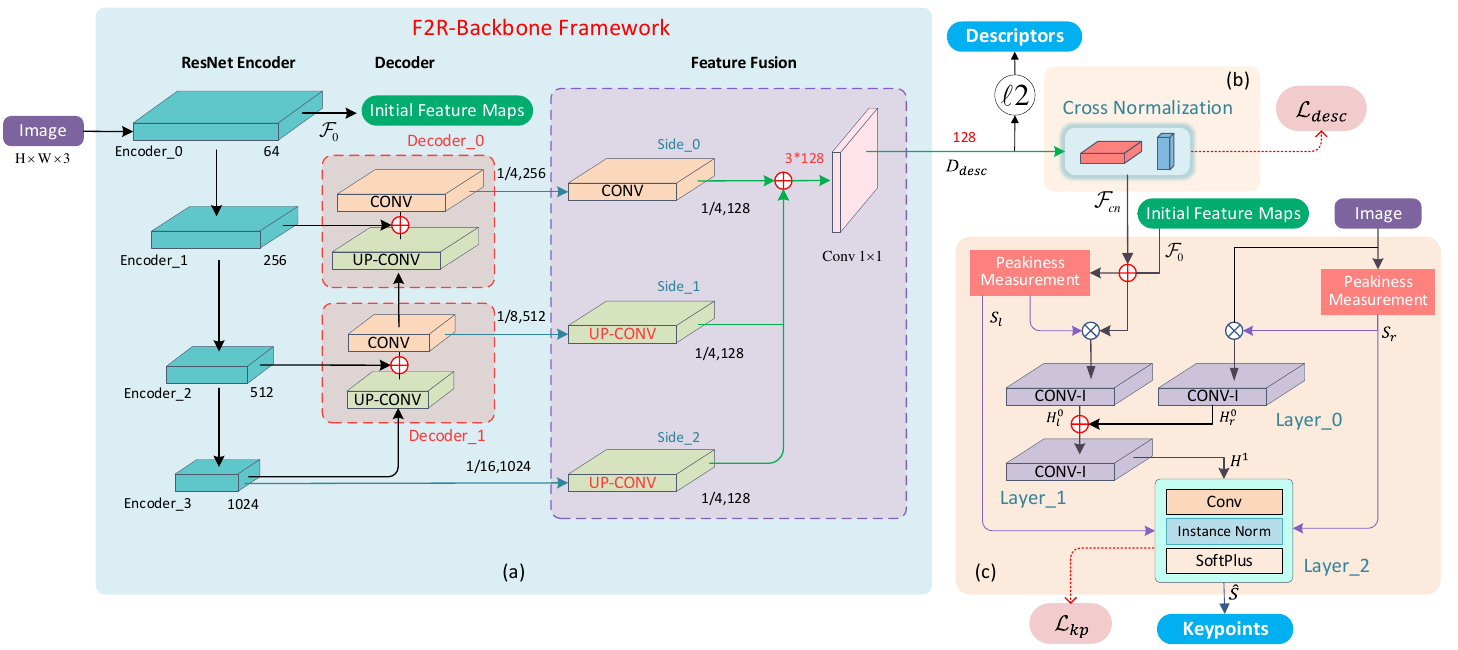} 
\caption{Systematical structure of proposed SCFeat: (a) description network, (b) shared coupling-bridge normalization, and (c) detection network.}
\label{fig-overview}
\end{figure*}

\begin{figure}[t]
\centering
\includegraphics[width=0.85\columnwidth]{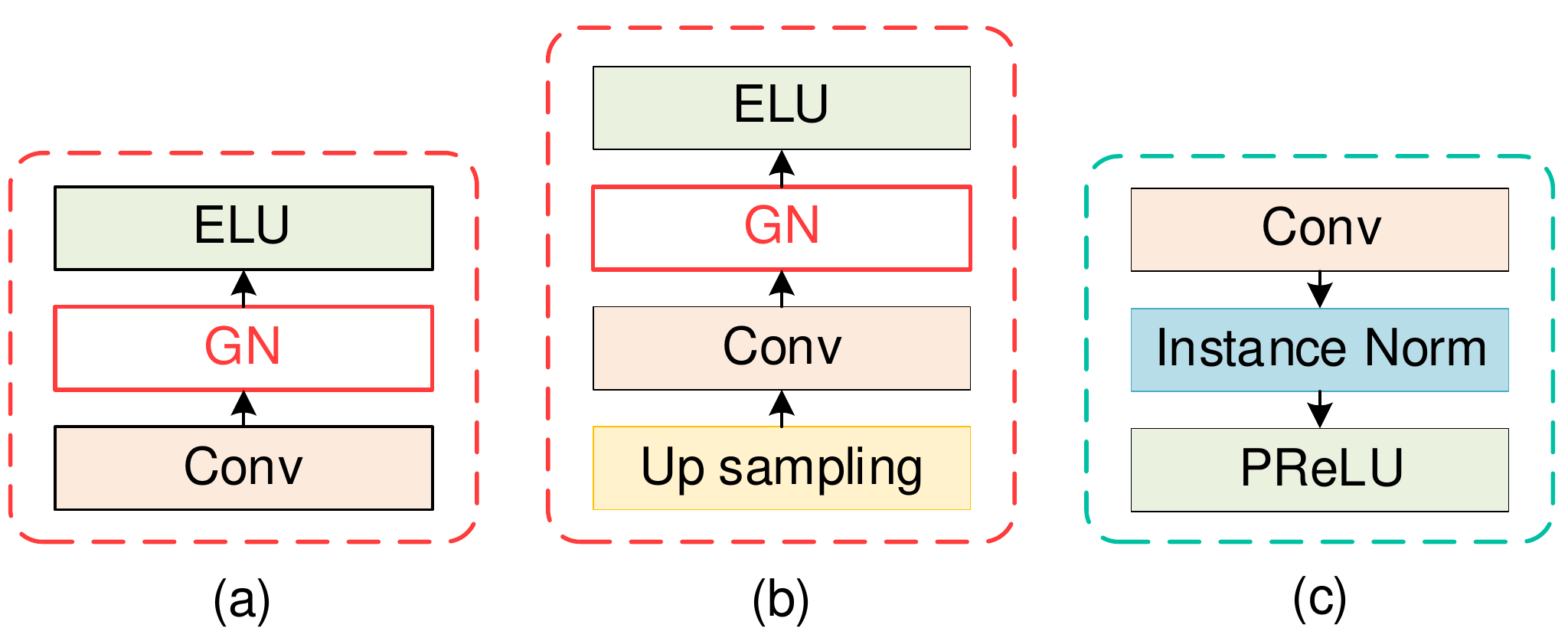} 
\caption{The details of three different convolution layers in Fig. \ref{fig-overview}: (a) the CONV layer with group normalization, (b) the UP-CONV layer with group normalization, and (c) the CONV-I layer with Instance Normalization.}
\label{fig3conv}
\end{figure}


For clarity, this paper uses the notations shown in TABLE \ref{table-symbols}. Our SCFeat is inspired by the PoSFeat in \cite{li2022decoupling}. It is powered mainly by these essential ingredients: i) the F2R-Backbone in Sec. \ref{description}, ii) the shared coupling-bridge normalization in Sec. \ref{CN}, iii) the detection network with peakiness measurement in Sec. \ref{detect}, and iv) the improved reward function on fundamental matrix error in Sec. \ref{loss} (2). Fig. \ref{fig-overview} shows the systematic structure of SCFeat.

Structurally, our SCFeat framework mainly consists of these two parts: description network and detection network. Seen from local feature extraction, it also belongs to a kind of describe-then-detect pipeline. Furthermore, in terms of structure design, it is a tightly-coupled scheme. From another point of view, it is also a decoupled-training one.







\subsection{Description Network} \label{description}


The basic architecture \& components of our SCFeat are illustrated in Fig. \ref{fig-overview} (a). Overall, it contains the three network modules: Resnet encoder group, decoder group and feature fusion group. Further, each group is made up of multiple function components.
Network modules, function components and their inter-connections compose the complete structure of description network for local feature descriptors.





\begin{itemize}
\item{
\emph{Four-stage ResNet Encoder Module}\\ This group includes Encoder$\_$0, Encoder$\_$1, Encoder$\_$2 and Encoder$\_$3. In details, we adopt the ResNet50 \cite{resnet} architecture, pre-trained on ImageNet and truncated after Encoder$\_$3 in our encoder group.
}
\item{
\emph{Two-stage Decoder Module}\\ 
This group contains two Decoder components of similar structures, $i.e.$, Decoder$\_$0 and Decoder$\_$1. Feature map is obtained by further convolutional layers along with up-sampling and skip-connections. Furthermore, each decoder consists of two convolution layers, $i.e.$, CONV and UP-CONV, respectively illustrated in Fig. \ref{fig3conv} (a) and (b). And both CONV and UP-CONV contain a $3\times3$ convolution sub-layer, a Group-Normalization (GN) sub-layer \cite{GN} and an activation sub-layer of ELU.
}
\item{
\emph{Feature Fusion Module}\\
This fusion module is mainly made up of 4 convolution layers, $i.e.$, a CONV layer, two UP-CONV layers and a general convolution layer (Conv 1$\times$1). It obtains three-side feature maps, respectively from Encoder$\_$3, Decoder$\_$1 and Decoder$\_$0, utilizing either CONV layer or UP-CONV layer. Then, it will fuse these side feature maps by the direct concatenation operation followed by a simple $1\times1$ convolution, to generate a group of dense feature maps. 
}
\end{itemize}

Seen on this description network architecture, we essentially propose a new strategy of \emph{Feature-Fusion-ResUNet Backbone} (F2R-Backbone) to enhance the quality of dense feature maps and stimulate the performance of shallow networks without more complex components and operations. By stacking, it could even afford us a deeper network of multi-scale features with higher computation \& memory costs. 

Finally, the dense descriptor maps $D_{desc}\in\mathbb{R}^{H\times W\times C}$, generated by description network, can be expressed by:
\begin{equation}\label{eq-ddm}
        D_{desc}=\mathcal{H}_{conv}(\mathcal{H}_{cat}([\mathcal{F}_{s0},\mathcal{F}_{s1},\mathcal{F}_{s2}]))
\end{equation}
where $\mathcal{H}_{cat}$ is a concatenate operation.  $\mathcal{H}_{conv}$ represents a $1\times1$ convolutional layer. $\mathcal{F}_{s0}$, $\mathcal{F}_{s1}$ and $\mathcal{F}_{s2}$ are three feature maps, simultaneously obtained respectively by Side$\_$0, Side$\_$1 and Side$\_$2, as shown in Fig. \ref{fig-overview} (a). 
Here, $\mathcal{H}(\cdot)$ can be a composite function of different operation layers such as convolution, ELU, and Instance Normalization (IN) \cite{IN}.



\subsection{Shared Coupling-Bridge Normalization}\label{CN}

As introduced in Sec. \ref{relatedworksB}, weakly-supervised loss cannot often distinguish the possible errors generated respectively by inaccurate descriptors and false keypoints. Moreover, the weakly-supervised loss only with camera-pose supervision cannot often powerfully distinguish this kind of descriptors without obvious discrimination. It usually makes the distinctiveness of generated descriptors lose in description network training. 
\begin{figure}[h]
\centering
\includegraphics[width=0.9\columnwidth]{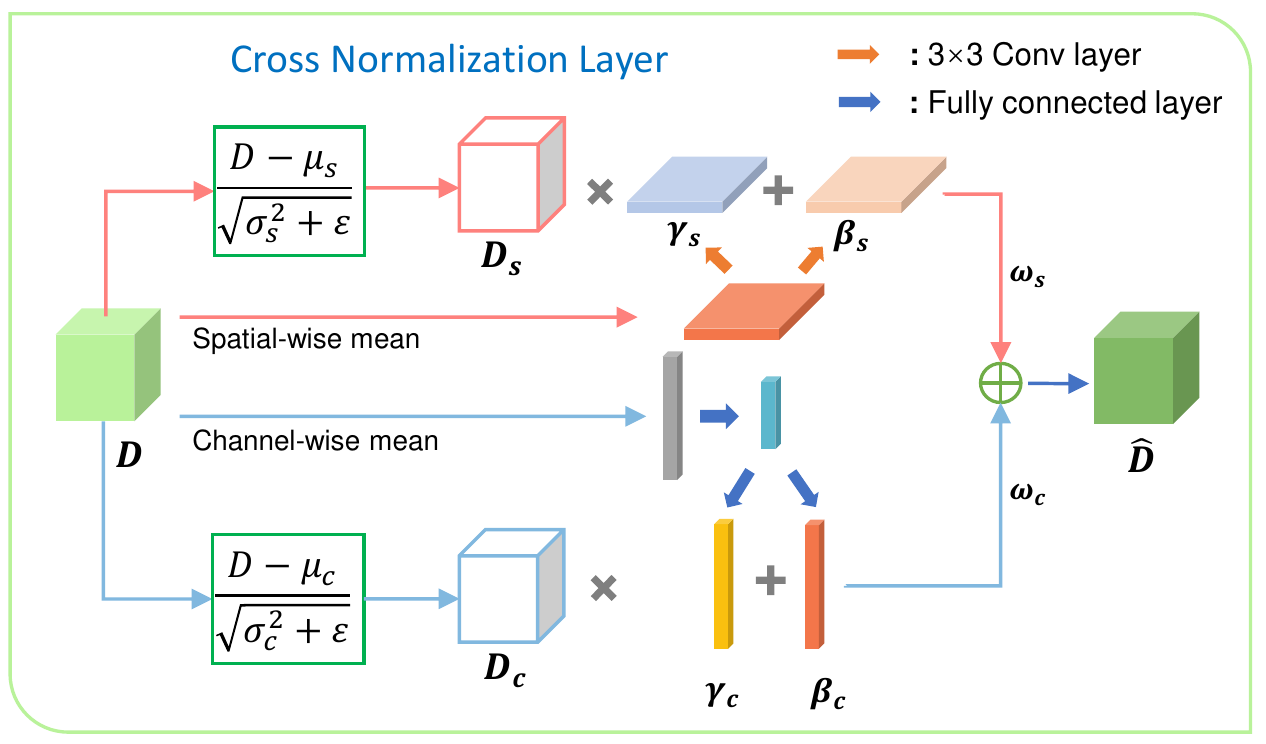}
\caption{Block diagram of the Cross Normalization Layer \cite{cndesc}.}
\label{fig_cnl}
\end{figure}


In order to promote the weakly-supervised local feature learning ability of our SCFeat,  
inspired by CNDesc \cite{cndesc}, we specially design a learnable cross normalization layer, principally illustrated in Fig. \ref{fig_cnl}, between description network and detection network. 
In terms of functionality, it exactly plays a role of shared coupling-bridge normalization.
In details, this normalization could be mathematically explained by
\begin{equation}\label{eq-cn}
        \hat{D}=(\gamma_s \times\frac{D-\mu_s}{\sqrt{\sigma_s^{2}+\varepsilon}}+\beta_s)\times w_s
        \oplus
        (\gamma_c \times\frac{D-\mu_c}{\sqrt{\sigma_c^{2}+\varepsilon}}+\beta_s)\times w_c
\end{equation}
where $\mu$ and $\sigma$ are the mean and standard deviation of a given channel and spatial dimension, respectively. Moreover, $\gamma$, $\beta$ and $w$ indicate a scale coefficient, a shifting parameter, and a weight coefficient for spatial or channel dimension to be learned, respectively. Besides, $\varepsilon$ represents an extra tiny constant to help preserve numerical stability.

\begin{figure}[h]
\centering
\includegraphics[width=0.9\columnwidth]{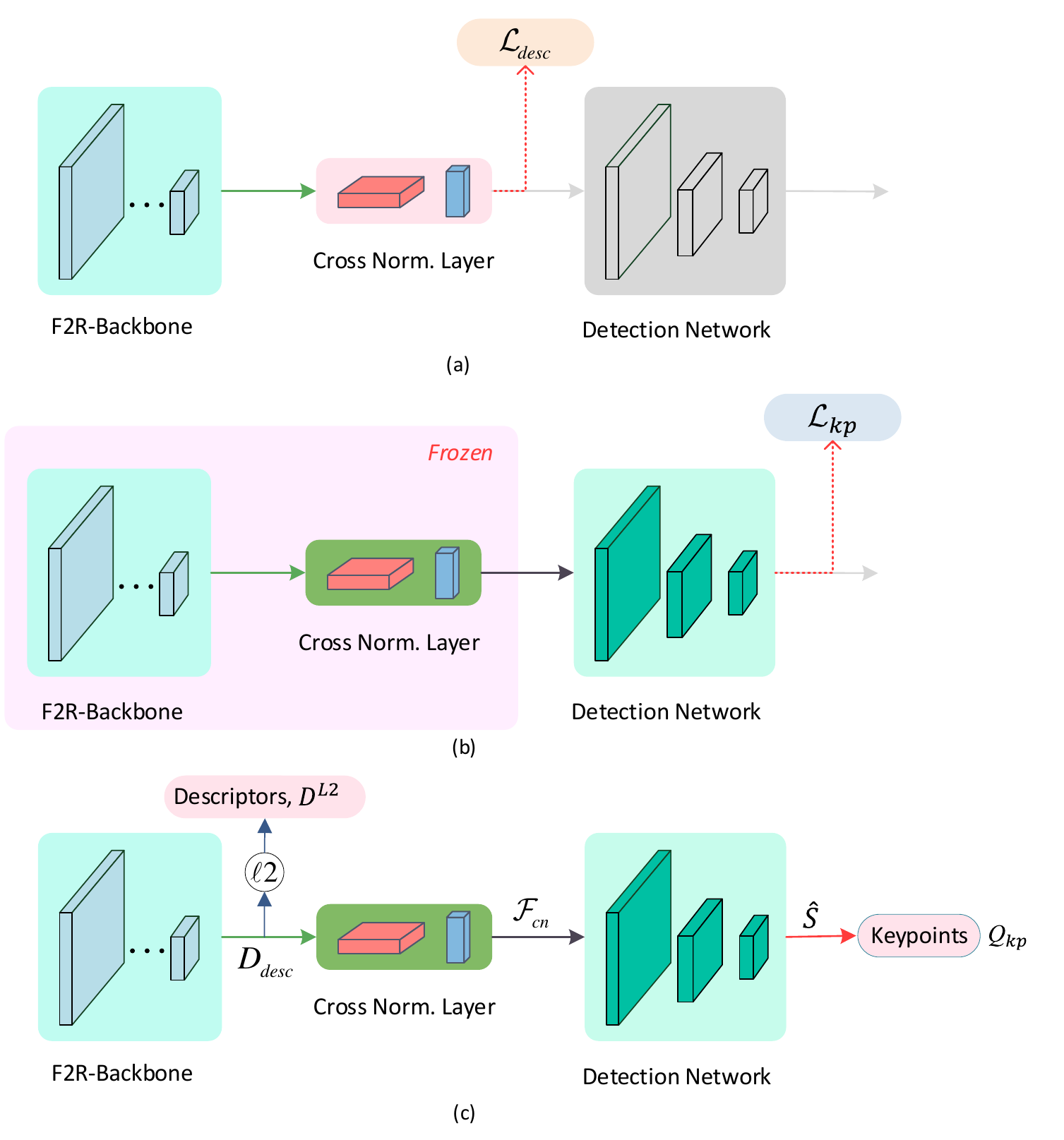} 
\caption{SCFeat Statuses: (a) F2R-Backbone training, (b) detection network training, and (c) local features (keypoints \& descriptors) extraction.}
\label{fig_cnstatus}
\end{figure}

In our SCFeat scheme, this coupling bridge of cross normalization will work as two different statuses: learnable and frozen, as illustrated comparatively in Fig. \ref{fig_cnstatus}. In the training of description network, this normalization layer is learnable, and the training loss, $\mathcal{L}_{desc}$ is computed by the output of this layer, as shown in Fig. \ref{fig_cnstatus} (a). In contrast, it will be frozen when training detection network, as shown in Fig. \ref{fig_cnstatus} (b). Moreover, when our SCFeat practically works to extract local features, this normalization layer still be in its station, merely the descriptors of local features are extracted ahead of this coupling normalization layer, illustrated in Fig. \ref{fig_cnstatus} (c).


For training detection network, original images and the dense feature maps from description network are taken as available inputs. And then, these inputs are aggregated together for feature detection. After doing this, we could train the detection network with only camera-pose supervision. However, the same dilemma as in description network training, also usually exists in detection network training. 
Therefore, for performance promotion, we consider preserving this cross-normalization layer (unlearnable, frozen with F2R-Backbone) when training detection network. 

From another angle, as shown in Fig. \ref{fig-overview}, such a cross normalization layer works exactly like a shared coupling bridge between description network and detection network in proposed SCFeat. 
Instead of the joint training of both description network and detection network, this bridge could transmit trained description parameters to the training of detection network, optimizing keypoint learning.







In details, the dense feature maps $\mathcal{F}_{cn}$, output from cross normalization layer, can be expressed by:
\begin{equation}\label{eq-dmap-cn}
        \mathcal{F}_{cn}=\mathcal{H}_{cn}(D_{desc})
\end{equation}
where $\mathcal{H}_{cn}$ presents an operation by cross normalization layer, and $D_{desc}$ has the same meaning as in Eqn. (\ref{eq-ddm}). 

In addition, in terms of pipeline, our shared coupling-bridge cross normalization could still maintain the important characteristics of \emph{describe-then-detect}.
Such a pre-trained cross normalization layer collects the training information of description network, and then it transmits that information to detection network training. 
Therefore, this scheme could increase the coupling of whole network, and enable our decoupled training to indirectly achieve the effects of joint optimization as same as possible. 


\subsection{Detection Network} \label{detect}
Overall, our detection network takes original images, the dense feature maps from cross normalization and the initial feature maps from Encoder\_0 as its detection inputs, and then aggregates them together for 
training and detecting keypoints, as illustrated in Fig. \ref{fig-overview} (c). 

\textbf{(1) Detection Head}

We design a three-layer architecture for the detection head and its functional components in our SCFeat:
\begin{itemize}
\item{The first layer, $i.e.$, Layer\_0, consists of two parallel convolution layers ($i.e$, CONV-I). 
Each CONV-I contains a sub-layer of general $3\times3$  convolution, an instance normalization sub-layer and an activation sub-layer of PReLU, as illustrated in Fig. \ref{fig3conv} (c).
This layer has two different inputs. One is the aggregation of initial feature maps ($\mathcal{F}_0$) and cross normalization output ($\mathcal{F}_{cn}$) with \emph{Peakiness Measurement} module. Similarly, the other is the aggregation of a given original image and its \emph{Peakiness Measurement} results.
 }
 
\item{The second layer, Layer\_1, only contains a single CONV-I layer. Its input is obtained by the up-sampling and skip-connection of Layer\_0.
}
\item{The last layer, Layer\_2 is made up of a $1\times1$ convolution sub-layer, an instance normalization sub-layer and an activation sub-layer of \emph{SoftPlus} activation.
}
\end{itemize}

\textbf{(2) Peakiness Score Measurement}

In order to make full use of the training information from shared coupling bridge for more accurate keypoints localization, inspired by \cite{D2net,aslfeat}, 
we specially design a weight term of peakiness measurement scores in this detection network. 
It scores a candidate interest point by regarding both spatial and channel-wise responses. In details, for each point $P(i,j)$ in the $k$-th dense feature map ${ D^{k}_{desc}}$ of $D_{desc}$, where $k=0, 1, 2,…, C-1$, its channel-wise peakiness score is computed by:
\begin{equation}\label{2}
\eta_{(i,j)}^k=\mathcal{H}_{softplus}\big(\boldsymbol{\rm y}_{(i,j)}^k-\frac{1}{C}\mathop{\sum}\limits_t\boldsymbol{\rm y}_{(i,j)}^{t}\big)
\end{equation}
where $\boldsymbol{\rm y}=D_{desc}$, 
$\mathcal{H}_{softplus}$ is a $\emph{SoftPlus}$ activation layer to transform a peakiness input into a non-negative one, and $t=0,1,...,C-1$. In the meanwhile, the local score of point $P(i,j)$ is generated by:
\begin{equation}\label{3}       \zeta_{(i,j)}^{k}=\mathcal{H}_{softplus}\big(\boldsymbol{\rm y}_{(i,j)}^{k}-\frac{1}{|\mathcal N(i,j)|}\mathop{\sum}\limits_{(i^{\prime},j^{\prime})\in \mathcal N(i,j)}{\boldsymbol{\rm y}_{(i^{\prime},j^{\prime})}^{k}}\big)
\end{equation}
where $\mathcal N(i,j)$ represents a group of neighboring pixels centered around $P(i,j)$, $e.g.$, the nine neighbors covered by a $3\times 3$ mask kernel. 

Moreover, to synthetically consider both $\eta_{(i,j)}^k$ and $\zeta_{(i,j)}^{k}$, we choose the maximum of their products on all feature channels to generate a final peakiness score map $S$ by:
\begin{equation}\label{4}
        S_{(i,j)}=\mathop{max}\limits_{k=0,1,...,C-1}(\zeta_{(i,j)}^{k}\times \eta_{(i,j)}^{k})
\end{equation}
where $S(i,j)$ is the maximal peakiness score of $P(i,j)$.

Therefore, based these definitions above, the output of \emph{Peakiness Measurement} could be briefly presented as $\mathcal{H}_{peak}(\boldsymbol{\rm y})=S$, where $\mathcal{H}_{peak}(\cdot)$ exactly represents a \emph{Peakiness Measurement} processing operation.

\textbf{(3) Input and Output Operation}

As described in Sec. \ref{detect} (1), 
this detection network takes original image $I_o \in \mathbb{R}^{H\times W \times 3}$, the $\mathcal{F}_{cn}$ from cross normalization layer and the $\mathcal{F}_{0}$ from Encoder\_0 as its three separate inputs. 

And then, it uses peakiness measurements $\mathcal{H}_{peak}$ to compute two peakiness score maps, generating $S_l \in \mathbb{R}^{\frac{H}{4} \times \frac{W}{4}}{}$ and $S_r \in \mathbb{R}^{H\times W}$ by
\begin{align}\label{eq-ps-det}
        S_l&=\mathcal{H}_{peak}(\mathcal{F}_{cn} \oplus \mathcal{F}_{0}) \\
        S_r&=\mathcal{H}_{peak}(I_o)
\end{align}
With these two maps of peakiness scores, the first layer outputs, $H^0_l\in \mathbb{R}^{\frac{H}{4} \times \frac{W}{4} \times C}$ and $H^0_r\in \mathbb{R}^{H\times W \times C}$, could be generated as
\begin{align}\label{eq-fo-det}
        H^0_l&=\mathcal{H}_{prelu}(\mathcal{H}_{in}(\mathcal{H}^{3\times3}_{conv}(S_l\otimes (\mathcal{F}_{cn} \oplus \mathcal{F}_{0})))) \\
        H^0_r&=\mathcal{H}_{prelu}(\mathcal{H}_{in}(\mathcal{H}^{3\times3}_{conv}(S_r\otimes I_{o})))
\end{align}
where $\mathcal{H}_{prelu}$ is a $\emph{PReLU}$ activation, $\mathcal{H}_{in}$ is an instance normalization, and $\mathcal{H}^{3\times3}_{conv}$ is a $3\times 3$ convolution. In Layer\_1, it aggregates the feature maps $H^0_l$ and $H^0_r$ from Layer\_0 to compute the output of Layer\_1, $H^1$ by
\begin{equation}\label{eq-o2}
        H^1=\mathcal{H}_{prelu}(\mathcal{H}_{in}(\mathcal{H}^{3\times3}_{conv}(\widetilde{H^0_l} \oplus H^0_r))))
\end{equation}
where $\widetilde{H^0_l}$ presents a bilinear interpolation on ${H^0_l}$. 

Similarly, the output of the last layer ($i.e.$, Layer\_2), $H^2$ could be further obtained by
\begin{equation}\label{eq-o3}
        H^2=\mathcal{H}_{softplus}(\mathcal{H}_{in}(\mathcal{H}^{1\times1}_{conv}(H^1)))
\end{equation} 
where $\mathcal{H}^{1\times1}_{conv}$ is a simple $1\times 1$ convolution. Finally, we use $S_l$ and $S_r$ as a weight coefficient for $H^2$ to generate the final score map $\widehat{\mathcal{S}}$ as
\begin{equation}\label{eq-s}
    \widehat{\mathcal{S}}=\widetilde{S_{l}} \otimes S_{r} \otimes H^2
\end{equation}
where $\widehat{\mathcal{S}} \in \mathbb{R}^{H\times W}$, and $\widetilde{S_l}$ still indicates a bilinear interpolation on peakiness score $S_l$.

\subsection{Weakly Supervised Local Feature Learning} \label{loss}

(1) \textbf{Epipolar Loss in Description Network Training}

Our training data consists of image pairs with relative camera poses. With only a weak supervision of camera poses, we conform to a $\emph{line-to-window}$ search strategy. Correspondingly, we adopt a loss function like the one as in \cite{li2022decoupling}:
\begin{equation}\label{1}
        \mathcal{L}_{desc}  =\frac{\sum_{i}{\frac{M_{i}}{\sigma(\boldsymbol{x}_{i})}\cdot \mathcal{L}_{epi}(\hat{\boldsymbol{y}_{i}},\boldsymbol{x}_{i})}}{\sum_{i}\frac{M_{i}}{\sigma(\boldsymbol{x}_{i})}}
\end{equation}
where $M_{i}$ is a binary mask (used to exclude the query points whose epipolar lines are not in given reference image) and $\sigma\left(\boldsymbol{x}_{i}\right)$ is the variance of probability distribution over local image patches. $\mathcal{L}_{epi}$ presents to compute the perpendicular distance between the predicted correspondence location $\hat{\boldsymbol{y}}_{i}$ and the ground-truth epipolar line $e(\boldsymbol{x}_{i})$, namely 
\begin{equation}\label{eq-epi}
        \mathcal{L}_{epi}(\hat{\boldsymbol{y}_{i}},\boldsymbol{x}_{i})={\rm dist}(\hat{\boldsymbol{y}}_{i},e(\boldsymbol{x}_{i}))
\end{equation}
where $\boldsymbol{x}_{i}$ represents the $i$-th given query point.


(2) \textbf{Reward Function in Detection Network Training}

Following a describe-then-detect pipeline with decoupled training, our description network and shared coupling bridge are frozen to produce the dense descriptor maps for feature detection after completing feature description learning. We adopt a similar strategy as in \cite{disk,li2022decoupling} to fulfil the training of feature detection network via policy gradient.

In only camera pose supervising, 
we firstly utilize a fundamental matrix error as an extra reward factor. And then, we propose a scheme of reward factor and epipolar distance to supervise detection network training.

In details, we firstly divide the score map of interest points $\widehat{S}$, generated by Eqn. (\ref{eq-s}) and $\widehat{S} \in \mathbb{R}^{H\times W}$, into a patch grid of size $g_k\times g_k$. In each grid cell, no more than one keypoint could be selected. Then, we use the keypoint probability distribution 
to select two candidate keypoint sets, $Q_1=\{\boldsymbol{x}_0,\boldsymbol{x}_1,...\}$ and $Q_2=\{\boldsymbol{y}_0,\boldsymbol{y}_1,...\}$ from an image pair $I_1$ and $I_2$, respectively. For $Q_1$ and $Q_2$, their two-way epipolar distances are computed by
\begin{gather}\label{eq-ed}
        E_{1}(\boldsymbol{y}_{i},\boldsymbol{x}_{i})={\rm dist}(\boldsymbol{y}_{i},e(\boldsymbol{x}_{i})) \\
        E_{2}(\boldsymbol{x}_{i},\boldsymbol{y}_{i})={\rm dist}(\boldsymbol{x}_{i},e(\boldsymbol{y}_{i}))
\end{gather}

Moreover, we introduce a distance threshold $\epsilon$ to determine a match matrix from the two confusion matrices of epipolar distances, $E_{1}$ and $E_{2}$, as follows:
\begin{equation}\label{6}
        P_{m}=(E_1<\epsilon) \& (E_2^{T}<\epsilon)
\end{equation}
where $P_m$ is an optimal match matrix for $Q_1$ and $Q_2$, symbol ``\&'' represents a logical-AND operator between two matrices.
Therefore, points $\boldsymbol{x}_i$ and $\boldsymbol{y}_j$ are regarded as a match pair if and only if $P_{m}(\boldsymbol{x}_i,\boldsymbol{y}_j)$ is $\emph{true}$. In our implementation, $\epsilon$ is often set to $2$. 

In addition, we will estimate a fundamental matrix error through both the fundamental matrix generated by an $\emph{Eight-Point}$ algorithm \cite{Hartley2004} and our iterative random re-sampling strategy. The pseudo-codes of proposed estimation algorithm have been listed in Algorithm \ref{alg:alg1}.

\begin{algorithm}[t]
\caption{Estimating Fundamental Matrix Error \protect\\
The parameters of the implemented model in this paper are defaulted as: $n_{iter}=100$}\label{alg:alg1}
\begin{algorithmic}[1]
\REQUIRE $n_{iter}$ number of iterations, $F$ is the ground-truth value of fundamental matrix from training data, $P_m$ is match matrix, $Q_{1}$ and $Q_{2}$ are candidate keypoints respectively select from an image pair $I_{1}$ and $I_{2}$.
\STATE $P_{int} \gets \textbf{toInt}(P_m)$
\STATE $V_{n} \gets \textbf{nonzero}(P_{int})$
\STATE $len \gets \textbf{length}(V_{n})$
\STATE $\textbf{assert } len>8$
\STATE $\textbf{initialize fundamental error vector } V_{\mathcal{L}_{fme}}$
\STATE $\textbf{for } k < n_{iter} \textbf{ do }$
\STATE \hspace{0.5cm}$ \textbf{select 8 indices randomly } N_{p} \subset \textbf{range}(0,len-1)  $
\STATE \hspace{0.5cm}$ \textbf{select 8 indices randomly } N_{p}^{\prime} \subset \textbf{range}(0,len-1)  $
\STATE \hspace{0.5cm}$ \mathbf{X}_{1} \gets Q_{1}[V_{n}[N_{p}[i]][1]] \textbf{ for } i = 0,...,7 $
\STATE \hspace{0.5cm}$ \mathbf{X}_{2} \gets Q_{2}[V_{n}[N_{p}^{\prime}[i]][2]] \textbf{ for } i = 0,...,7 $
\STATE \hspace{0.5cm}$ \hat{F} \gets \mathcal{H}_{eight}(\mathbf{X}_{1}, \mathbf{X}_{2}) $
\STATE \hspace{0.5cm}$ V_{\mathcal{L}_{fme}}[k] \gets \mathcal{H}_{smooth\_\ell 1}(\hat{F}, F) $
\STATE \textbf{end for}
\RETURN \textbf{min}$(V_{\mathcal{L}_{fme}})$
\end{algorithmic}
\label{alg1}
\end{algorithm}

The $\emph{Eight-Point}$ algorithm, presented as $\mathcal{H}_{eight}$, could estimate the fundamental matrix, $\hat{F}\in \mathbb{R}^{3\times 3}$ on two given images of a same scene, although not knowing the extrinsic or intrinsic parameters of camera poses. It assumes that a set of at least 8 pairs of corresponding points between two images is available. In practice, we often have thousands of candidate correspondences between two images, and in the meanwhile, there possibly exist a lot of noisy measurements. 

In order to overcome these problems, we propose a method of iterative random re-sampling with \emph{\rm Smooth L1} distance ($\mathcal{H}_{smooth\_\ell 1}$). Its optimization object is to minimize a fundamental matrix error, namely  
\begin{equation}\label{10}
        \mathop{\min} \mathcal{L}_{fme}=
        \mathcal{H}_{smooth\_\ell 1}(\hat{F},F)
\end{equation}
where $\mathcal{L}_{fme}$ indicates a loss of fundamental matrix, $\hat{F}$ is the optimal estimation of fundamental matrix in multiple iterations, 
and $F$ is the $\emph{Ground Truth}$ of fundamental matrix. 
Our iterative scheme could promote the robustness of local feature extraction system by producing uncertainty through random re-sampling. In fact, seen from another viewpoint, the fundamental matrix error estimated by our proposed algorithm could also reflect current matching error. Namely, we can evaluate the quality of optimal match matrix $P_{m}$ through such a fundamental matrix error.

Furthermore, our new reward function based on fundamental matrix error in our scheme is defined as:
\begin{multline}\label{11}
        r(\boldsymbol{x}_i,\boldsymbol{y}_j)=(1-tanh(\mathcal{L}_{fme}))P_{m}(\boldsymbol{x}_i,\boldsymbol{y}_j)-\\tanh(\mathcal{L}_{fme})(\neg P_{m}(\boldsymbol{x}_i,\boldsymbol{y}_j))
\end{multline}
Therefore, as defined in \cite{li2022decoupling}, we could finally redefine the completed loss function, $\mathcal{L}_{kp}$ as
\begin{multline}\label{12}
        \mathcal{L}_{kp}=\frac{1}{|Q_1|+|Q_2|}(\mathop{\sum}\limits_{\boldsymbol{x}_i,\boldsymbol{y}_j}{r(\boldsymbol{x}_{i},\boldsymbol{y}_{j})\log {(P_{kp}(\boldsymbol{x}_i)P_{kp}(\boldsymbol{y}_i))}} +\\
        \lambda_{reg}\mathcal{L}_{fme}(\mathop{\sum}\limits_{\boldsymbol{x}_i}{\log{P_{kp}(\boldsymbol{x}_i)}}+\mathop{\sum}\limits_{\boldsymbol{y}_j}{\log{P_{kp}(\boldsymbol{y}_j)}})),
\end{multline}
where $\lambda_{reg}\mathcal{L}_{fme}$ is a new regularization penalty, $P_{kp}$ is the keypoint probability distribution, besides all other expressions represent the same meanings as before.

\subsection{Local Feature Extraction}

The final extraction scheme of keypoints and its descriptors is briefly illustrated by Fig. \ref{fig_cnstatus} (c). 
It is easy to observe that final descriptors are generated by $D_{desc}$, $i.e.$, the direct output of description network, and in the meanwhile keypoints are extracted from the output of detection network.

In details, we only select top-$K$ candidate points 
as the final output of keypoints, according to the detection scores obtained by Eqn. (\ref{eq-s}), empirically discarding those candidates with low scores. Moreover, a non-maximum suppression (NMS) \cite{NMS} is also applied to further remove some redundant detection points too close to each other in spatial positions. 


 Besides, we additionally utilize a channel-wise L2 normalization to further optimize the original dense descriptor maps $D_{desc}$, output from description network and generated by Eqn. (\ref{eq-ddm}),
 aiming to refine better local feature descriptors. 
 Mathematically, this normalization is expressed by
\begin{equation}\label{eq-l2}
        d_{L2}=\frac{d}{\sqrt{\sum_{i=0}^{n-1}{d_{i}^{2}}}}
\end{equation}
where $d$ is a local descriptor vector to be normalized, $d_i$ represents the $i$-th component of $d$, and $n$ is the size of given $d$.

Finally, the normalized local descriptors, $D^{L2}\in\mathbb{R}^{K\times C}$, could be obtained by
\begin{equation}\label{eq-dd}
        D^{L2}=\mathcal{H}_{\ell 2}(\mathcal{H}_{grid\_sample}(D_{desc},Q_{kp}))
\end{equation}
where $\mathcal{H}_{\ell 2}$ indicates an L2 normalization and $\mathcal{H}_{grid\_sample}$ represents an operation of grid sampling. In details, grid sampling means an operation that computes an output according to given input and the pixel locations from grid, under a given input and a flow-field grid. Here, $Q_{kp}=\{\boldsymbol{x}_{1},\boldsymbol{x}_{2},...,\boldsymbol{x}_{K}\}$ is a set of keypoints determined via the score map $\widehat{\mathcal{S}}$, computed by Eqn. (\ref{eq-s}), under the limitation of scores and the maximal number of expected keypoints.

\section{Experiments and Comparisons}

To evaluate the performance and advantages of our sparse local feature extraction scheme, $i.e.$, SCFeat, we conduct three groups of experiments, including: i) the local feature matching on dataset HPatches \cite{HPatches}, ii) the visual localization on dataset Aachen Day-Night \cite{aachen} and iii) the 3D reconstruction with ETH local feature benchmark \cite{ETH}.





\subsection{Experiment Settings}


\textbf{(1) Available Datasets}

We use MegaDepth dataset 
\cite{MegaDepthLi18} for training, which consists of 196 different scenes reconstructed from over 1M internet photos using COLMAP \cite{schoenberger2016sfm,schoenberger2016mvs}. Totally, 130 out of 196 scenes are used for training, and we train our system on image pairs using provided camera poses and intrinsics.  

To validate the performance of our proposed SCFeat, we conduct experiments on the following three datasets respectively for three different task scenarios:
\begin{itemize}
\item{
The \emph{HPathces}\cite{HPatches}.
The most intuitive evaluation rule of local features is how accurately they can be matched between image pairs. This dataset is the most popular benchmark for image matching. It consists of 116 sequences of 6 images (totally 696 images), with 59 viewpoint variations and 57 illumination variations. 
}
\item{
The \emph{Aachen Day-Night v1.1}\cite{aachen}.
This dataset contains 6,697 reference images and 1,015 query images (824 day-time images, 191 night-time ones). All query images are taken by mobile phone cameras. This benchmark is challenging because it needs to match the image pairs between day and night. We adopt the official visual localization pipeline\footnote{https://github.com/ahojnnes/local-feature-evaluation.git} for the local feature challenge of workshop on long-term visual localization under changing conditions. This challenge only evaluates the pose of night-time query images.
}
\item{
The \emph{ETH dataset }\cite{ETH}.
The effectiveness of our local features is evaluated by the context of 3D reconstruction on the local feature benchmark of $\emph{ETH}$. In details, three medium-scale sub-datasets with a significant number of different cameras and conditions are used for experiments, including \emph{Madrid Metropolis} (1,344 images), \emph{Gendarmenmarkt} (1,463 images) and \emph{Tower of London} (1,576 images).
}

\end{itemize}

\textbf{(2) Training Setup}

All networks are trained by an Adam optimizer. In description network training, learning rate is set to $10^{-4}$ and batch size is set to 2. The description network of F2R-Backbone is trained for $100$ thousand iterations. In detection network training, the learning rate is set to $10^{-3}$ and batch size is set to 2. And detection network is trained only for $1$ thousand iterations. All training is conducted in the \emph{Pytorch} framework on a single NVIDIA RTX3060 GPU. 

\begin{figure*}[h]
\centering
\includegraphics[width=0.75\textwidth]{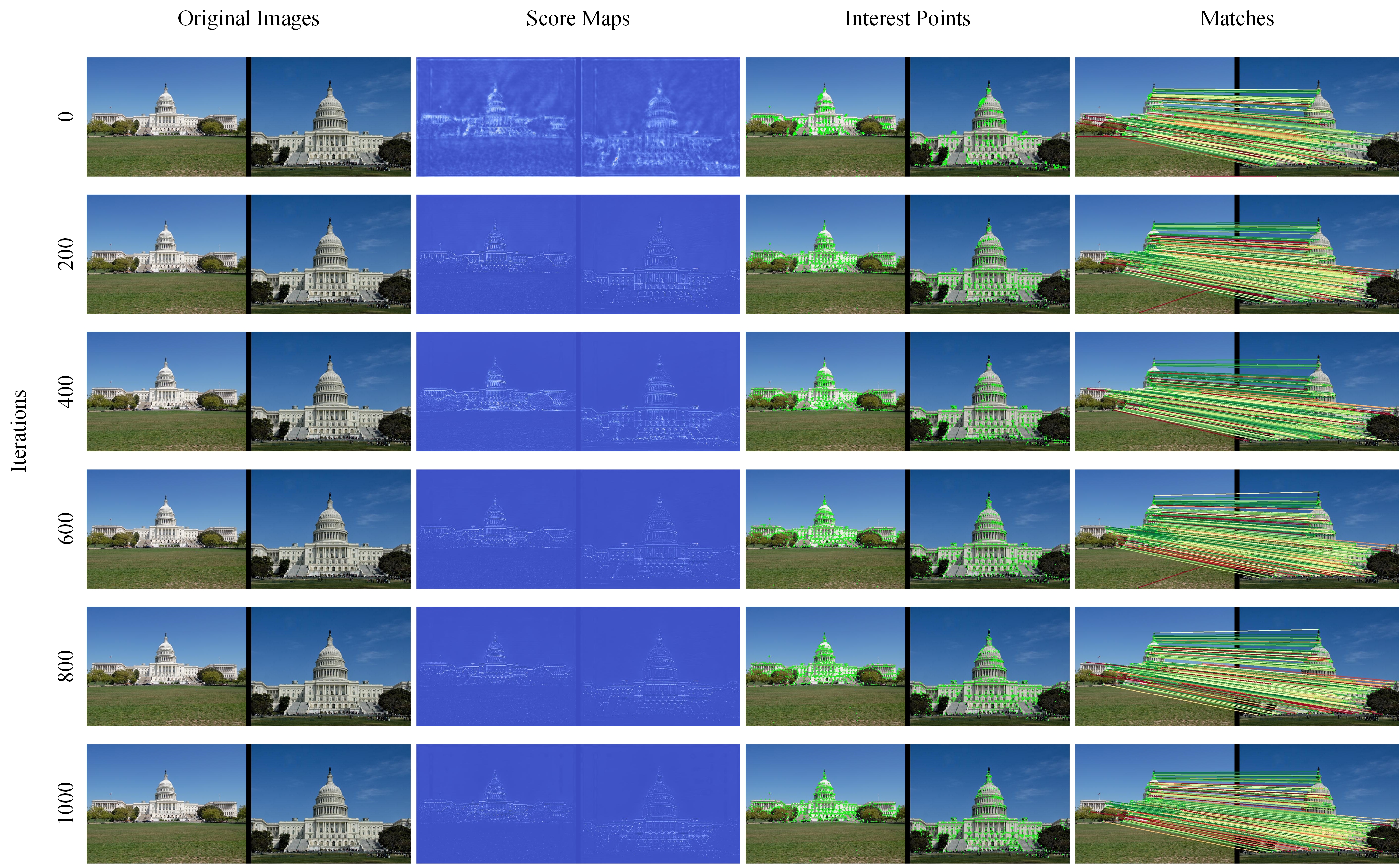}
\caption{Visualization of detection network training for our SCFeat. The maximum interest point number is limited to $2k$ and the maximum iteration is limited to $1000$.}
\label{fig30}
\end{figure*}

During inference, we use a mutual nearest neighbor matcher for matching. Moreover, in order to exhibit the efficiency of network training, Fig. \ref{fig30} shows a part of visualization results in detection network training for our SCFeat.

\subsection{Local Feature Matching}\label{sec-fm}

\textbf{(1) Comparisons on Image Matching}

The dataset used in this part is \emph{HPathches} \cite{HPatches}. It contains $116$ scenes, and in each scene the first image is taken as a reference while the others in a sequence are used to form the image pairs with increasing difficulty. Following \cite{D2net,li2022decoupling}, we exclude 8 high-resolution sequences, retaining $52$ and $56$ sequences with illumination variations and viewpoint variations, respectively.

The mean matching accuracy (MMA) with a threshold ranging from 1 to 10 as a primary evaluation metric. Its score is defined by the average percentage of correct matches for image pairs under a certain pixel error threshold. As in \cite{li2022decoupling}, we also use a weight sum of MMA at different thresholds for overall evaluation.

\begin{equation}\label{eq-mma}
        {\rm MMAscore}  =\frac{\sum_{thr=1}^{10}(2-0.1\times thr)\times {\rm MMA}@thr}{\sum_{thr=1}^{10}(2-0.1\times thr)}
\end{equation}
where ``$thr$'' exactly represents a given threshold, and ``${\rm MMA}@thr$" indicates the MMA score under a given threshold.

Moreover, for the non-maximum suppression (NMS) in our SCFeat, size $N$ is set to $1\times 1$ and the initial number for the maximal interest points is set to $8k$. Besides, we test SCFeat without NMS and maximum interest point number $20k$ ($i.e.$, Ours \textit{w/o NMS} in TABLE \ref{table1}). Keypoints with scores smaller than 1.0 in score maps are filtered out ($i.e.$, $\epsilon_{s}=1.0$, please see it in TABLE \ref{table-symbols}). 

We compare our SCFeat schemes with several recent baselines, including the Hessian affine detector with Root SIFT descriptor (Hes. Aff. + Root-SIFT) \cite{HesAffandSIFT}, the HesAffNet regions \cite{HesAffNet} with HardNet++ descriptors \cite{HardNetpp} (HAN + HN++), the SIFT detector with ContextDesc descriptor \cite{ContextDesc} (SIFT + ContextDesc), D2-Net \cite{D2net}, R2D2 \cite{R2D2}, ASLFeat \cite{aslfeat}, the DISK version \cite{disk} with weakly supervised learning  (DISK-W), DELF \cite{DELF}, the SIFT detector with CAPS descriptor \cite{caps} (SIFT + CAPS), SuperPoint \cite{SuperPoint} detector with CNDesc \cite{cndesc} descriptor, as well as PoSFeat \cite{li2022decoupling}.

\begin{table*}[h]
\renewcommand\arraystretch{1.5}
\centering
\caption{The Comparisons of different methods on HPatches, with the MMAscore calculated under the conditions in Fig. \ref{fig6}.}
{
\begin{threeparttable}
\begin{tabular}{l|cc|ccc}
    \hline
    \multirow{2}{*}{Methods} & Num. & Num. & \multicolumn{3}{c}{MMAscore (\%)}\\
    \cline{4-6}
            & Feature & Match & Viewpoint & Illumination & Overall \\
    \hline
    \hline
    Hes. Aff. + Root-SIFT\tnote{1} \ \cite{HesAffandSIFT}  & 6710.1 & 2851.7 & 62.1 & 54.4 & 58.4 \\
    HAN \cite{HesAffNet} + HN++\tnote{1} \ \cite{HardNetpp}             & 3860.8 & 1960.0 & 63.3 & 63.4 & 63.3 \\
    SIFT + ContextDesc\tnote{1} \ \cite{ContextDesc}     & 4066.7 & 1744.3 & 65.7 & 61.3 & 63.6 \\
    D2-Net\tnote{1} \ \cite{D2net}                 & 2994.1 & 1182.6 & 44.0 & 60.5 & 51.9 \\
    R2D2\tnote{1} \ \cite{R2D2}                   & 4996.0 & 1850.5 & 66.5 & 72.7 & 69.5 \\
    ASLFeat\tnote{1} \ \cite{aslfeat}                & 4013.5 & 2009.8 & 68.7 & 79.5 & 73.9 \\
    DELF\tnote{1} \ \cite{DELF}                   & 4590.0 & 1940.3 & 26.2 & \textbf{90.3} & 57.1 \\
    SIFT + CAPS\tnote{1} \ \cite{caps}            & 4386.7 & 1450.4 & 63.9 & 76.4 & 69.9 \\
    DISK-W\tnote{1} \ \cite{disk}                 & 6759.5 & 3976.4 & 64.9 & 80.3 & 72.3 \\
    SuperPoint + CNDesc\tnote{2} \ \cite{cndesc} & 1453.5 & 787.9 & 72.0 & 74.8 & 73.4 \\
    PoSFeat\tnote{1} \ \cite{li2022decoupling}                 & 8192.0 & 4275.3 & 72.8 & 82.6 & 77.5 \\
    \hdashline
    Ours      & 4875.1 & 2482.2 & \underline{73.4} & 84.6 & \underline{78.8} \\
    Ours \textit{w/o NMS}          & 14876.1 & 3834.0 & \textbf{75.8} & \underline{88.7} & \textbf{82.0} \\
    \hline
\end{tabular}
\begin{tablenotes}
\footnotesize
\item[1] MMAscores reported at \cite{li2022decoupling}.
\item[2] MMAscores obtained by ourselves, running the code with its default configuration.
\end{tablenotes}
\end{threeparttable}
}
\label{table1}
\end{table*}

\begin{figure*}[h]
\centering
\includegraphics[width=1.8\columnwidth]{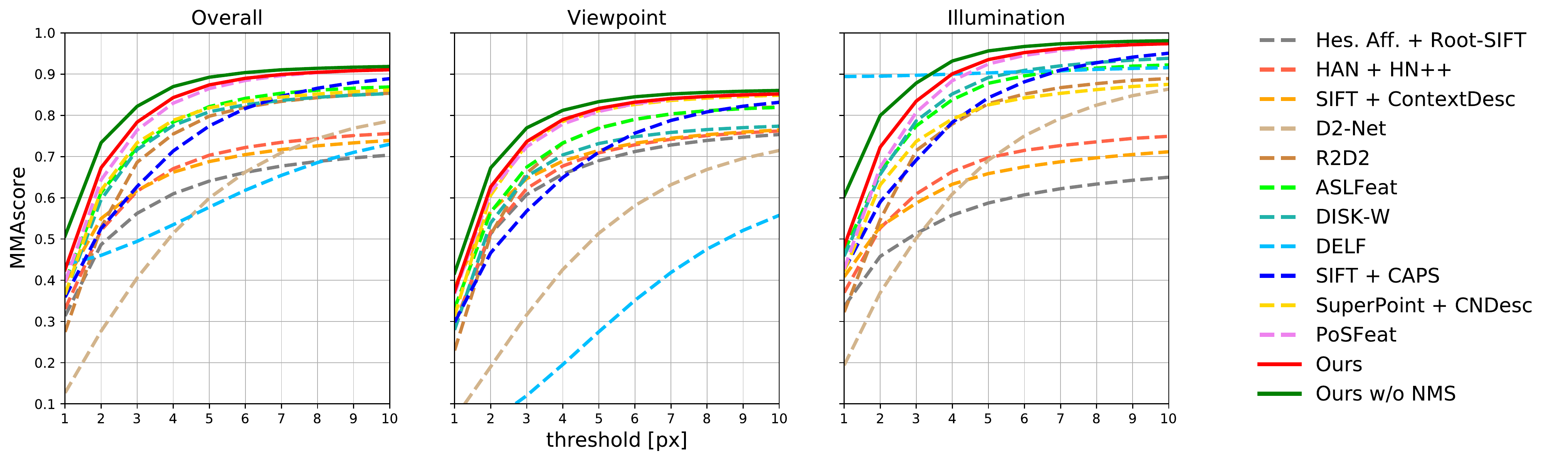}
\caption{Results achieved on HPatches dataset. Mean matching accuracy (MMA), evaluated at different error thresholds, is illustrated.}
\label{fig6}
\end{figure*}

Both Fig. \ref{fig6} and TABLE \ref{table1} exhibit our comparison results.
It is easy to observe that our SCFeat could often achieve the best overall performance. Under same conditions, SCFeat outperforms PoSFeat on both illumination (84.6 vs. 82.6) and viewpoint changes (73.4 vs. 72.8). Meanwhile, it achieves higher overall MMA than PoSFeat (78.8 vs. 77.5). Moreover, we visualize score map (left) and keypoints (right) in Fig. \ref{fig2}. 
Compared to PoSFeat, the features extracted by our SCFeat are more accurate. 

Furthermore, ``Ours \textit{w/o NMS}" achieves the best performance (82.0) in all comparisons on ``Overall''. It could prove that the features extracted by SCFeat have better comprehensive accuracy and discrimination. We also find that DELF is the best one only on ``illumination''. However, our SCFeat could significantly outperform it on ``viewpoint'' (75.8 vs. 26.2) and ``overall'' (82.0 vs. 57.1).

\textbf{(2) Ablation Study}

\begin{table*}[h]
\centering
\renewcommand\arraystretch{1.5}
\caption{Ablation Comparisons: the MMAscore results achieved by different combination schemes on HPatches \cite{HPatches}. 
}
{
\begin{tabular}{l|cc|ccc}
    \hline
     \multirow{2}{*}{Methods} & Num. & Num. & \multicolumn{3}{c}{MMAscore (\%)}\\
    \cline{4-6}
            & Feature & Match & Viewpoint & Illumination & Overall  \\
    \hline
    \hline
    Ours \textit{w/o error factor}          & 4285.3 & 2099.7 & 71.9 & 84.2 & 77.8  \\
    Ours \textit{w/ error factor}               & 4875.1 & 2264.4 & \textbf{73.4} & \textbf{84.6} & \textbf{78.8} \\
    \hline
    \hline
    SuperPoint + CAPS \cite{caps}     & 1562.6 & 870.5 & \textbf{69.0} & 76.7 & \textbf{72.7} \\
    SuperPoint + CAPS + CN              & 1562.6 & 840.5 & 62.9 & 75.5 & 69.0 \\
    \hdashline
    SuperPoint + CAPS + CN \textit{w/ pruning}  & 1562.6 & 702.8 & \underline{64.9} & \textbf{76.8} & \underline{70.6} \\
    \hline
    \hline
    Ours \textit{w/o pruning}       & 4875.1 & 2340.0 & 72.5 & \textbf{84.9} & 78.5 \\
    Ours \textit{w/o cross norm.}               & 695.9 & 162.2 & 3.8 & 73.5 & 37.3 \\
    \hdashline
    Ours    & 4875.1 & 2264.4 & \textbf{73.4} & \underline{84.6} & \textbf{78.8} \\
    \hline
    \hline
    SuperPoint \textit{w/} CAPS \cite{caps}       & 1562.6 & 870.5 & 69.0 & 76.7 & 72.7 \\
    SuperPoint \textit{w/} CNDesc \cite{cndesc}              & 1453.5 & 787.9 & 72.0 & 74.8 & 73.4 \\
    \hdashline
    SuperPoint \textit{w/} F2R-Backbone \textit{\& $\ell$2 norm.}  & 1552.8 & 925.0 & 69.9 & 78.5 & \underline{74.0} \\
    SuperPoint \textit{w/} F2R-Backbone \textit{\& fusion norm.}   & 1552.8 & 805.5 & \textbf{74.9} & \textbf{82.7} & \textbf{78.7} \\
    \hline
\end{tabular}
}
\label{table4}
\end{table*}

To analyze the effectiveness of different SCFeat schemes, we conduct a group of ablation experiments, including our F2R-Backbone, proposed reward factor, and shared normalization coupling bridge. 
Our experiments still follow the evaluation protocol in Sec. \ref{sec-fm} (1) and compare MMA over all the image pairs in HPatches. TABLE \ref{table4} exhibits our numerical results of ablations.



In this table, ``Ours \textit{w/ error factor}" indicates the SCFeat version with our new reward function and ``Ours \textit{w/o error factor}" represents the one with only the epipolar distance reward in PoSFeat \cite{li2022decoupling} (Both are trained by $1k$ iterations).

In addition, ``SuperPoint + CAPS" is the SuperPoint \cite{SuperPoint} detector with CAPS \cite{caps} descriptor, ``SuperPoint + CAPS + CN" is the SuperPoint detector with CAPS descriptor and cross normalization,
and ``SuperPoint + CAPS + CN \textit{w/ pruning}" means that cross normalization is used in the training of CAPS and then removed in descriptors extraction.

``Ours \textit{w/o pruning}" indicates that we use the dense descriptor maps output from coupling-bridge cross normalization for descriptors extraction. We also test the detection network without cross normalization, $i.e.$, ``Ours \textit{w/o cross norm.}", and the full SCFeat is represented by ``Ours".

``SuperPoint \textit{w/} CAPS" is the SuperPoint detector with CAPS descriptor, ``SuperPoint \textit{w/} CNDesc" is the SuperPoint detector with CNDesc \cite{cndesc} descriptor, ``SuperPoint \textit{w/} F2R-Backbone \textit{\& $\ell$2 norm.}" uses SuperPoint detector and the F2R-Backbone descriptor with L2 normalization, and ``SuperPoint \textit{w/} F2R-Backbone \textit{\& fusion norm.}" uses SuperPoint detector and the F2R-Backbone descriptor with fusion normalization. In this part, fusion normalization is a new scheme, temporarily designed here, to further investigate the effect of $L_p$ normalization on local descriptors. It is defined by
\begin{equation}\label{eq-fusion-norm}
        \hat{D}=\frac{1}{\sum_{i=1}^{m}i}\sum_{p=1}^{m}{k_{p}\frac{d}{||d||_{p}}}
\end{equation}
where $||d||_{p}$ presents the $L_{p}$ norm of $d$, $||d||_{p}=({\sum_{i=1}^{n} d_{i}^{p}})^{\frac{1}{p}}$, $k_{p}$ is a weight coefficient and $m\in \mathbb{N}^{+}$.
When $m=2$, $k_1=0$ and $k_2=3$, fusion normalization and L2 normalization are equivalent. In ``SuperPoint \textit{w/} F2R-Backbone \textit{\& fusion norm.}", it sets $m=4$, $k_1=1$, $k_2=2$, $k_3=3$ and $k_4=4$.


i) \emph{New Reward Factor}

In the first of TABLE \ref{table4}, we could see that our new reward function obviously improves MMA performance, due to making full use of such geometry information in camera poses as epipolar distance and fundamental matrix, as well as iterative random re-sampling. For example, our new reward function promotes MMAscore from 77.8\% to 78.8\%, gaining an obvious improvement of 1\% on ``Overall''. 

ii) \emph{Cross Normalization in Descriptor Learning}

In the second part of
TABLE \ref{table4}, we observe that ``SuperPoint + CAPS'' is the best one (72.9\%), although 
``SuperPoint + CAPS + CN \textit{w/ pruning}" (70.6\%) outperforms ``SuperPoint + CAPS + CN" (69.0\%) on ``Overall''. This group of comparisons implies that cross normalization is often inefficient for \emph{CAPS} method. Simply adding a cross normalization layer cannot promote the performance of description network in weakly supervised learning.

One of possible reasons, we speculate, is that this cross normalization usually ignores the details of dense feature maps in weakly supervised learning, declining description ability. Therefore, removing cross normalization layer after training could effectively improve this situation. 

iii) \emph{Shared Normalization Coupling Bridge} 

In the third part of TABLE \ref{table4}, it could be found that ``Ours \textit{w/o cross norm.}" achieves the worst performance. For example, it only reaches 37.3\% on ``Overall'', much lower than the 78.8\% by ``Ours''. Especially, it only gains 3.8\% on ``Viewpoint''. This group of companions infers that the Shared Normalization Coupling Bridge designed by us could be effective and necessary to our SCFeat scheme. 


iv) \emph{The Impact of F2R-Backbone}

In the last part of TABLE \ref{table4}, it is easy to observe that ``SuperPoint \textit{w/} F2R-Backbone \textit{\& fusion norm.}" is always the best one on ``Viewpoint'', ``Illumination'' and even ``Overall''. 
For example, 
``SuperPoint \textit{w/} F2R-Backbone \textit{\& $\ell$2 norm.}" obtains an MMAscore of 74.0\% on ``Overall'', 1.3\% higher than the 72.7\% of ``SuperPoint \textit{w/} CAPS", and 0.6\% higher than the 73.4\% of ``SuperPoint \textit{w/} CNDesc".
Furthermore, ``SuperPoint \textit{w/} F2R-Backbone \textit{\& fusion norm.}" even achieves 78.7\% on ``Overall'', 4.7\% higher 
than the 74.0\% of ``SuperPoint \textit{w/} F2R-Backbone \textit{\& $\ell$2 norm.}".

On hand, since these four schemes share the same detector of SuperPoint, this group of comparisons implies that our descriptor, $i.e.$, F2R-Backbone with \textit{fusion norm.}, is efficient and superior to both CAPS and CNDesc for SuperPoint.
On the other hand, since the last two share a same pair of descriptor \& detector, this group of comparisons implies that \textit{fusion norm.} could usually outperform \textit{$\ell$2 norm.} for such a pair of local feature extraction networks. 

One of the most possible reasons is that the fusion normalization could keep more information from dense descriptor maps and improve the repeatability of matching between different descriptors.

\subsection{Visual Localization} \label{vl}

In this section, we evaluate our SCFeat in the context of visual localization on Aachen Day-Night v1.1 \cite{aachen}. The evaluation pipeline in our experiments is provided by visual localization benchmark. It takes custom local features as input, and then relies on COLMAP in \cite{schoenberger2016sfm,schoenberger2016mvs} for image registration. After doing that, it computes the percentages of successfully localized images within the three error tolerances: \emph{high-precision} $(0.25m, 2^{\circ})$, \emph{medium-precision} $(0.5m, 5^{\circ})$ and \emph{coarse-precision} $(5m, 10^{\circ})$.

Considering the resolutions of images, NMS size $N$ is set to 5$\times$5 for reference (database) images and 3$\times$3 for query images. Meanwhile, $\epsilon_{s}$ is set to 1.0 and 0.5 for reference images and query images, respectively. The maximum number of keypoints is limited to $20k$. 
In experiments, our SCFeat is compared with several recent baselines of extractor methods: R2D2 \cite{R2D2}, ASLFeat \cite{aslfeat}, the SuperPoint detector \cite{SuperPoint} with LISRD descriptor \cite{LISRD} and AdaLAM outlier detection \cite{cavalli2020handcrafted} (LISRD+SuperPoint+AdaLAM), the SuperPoint detector with CNDesc descriptor \cite{cndesc} (CNDesc+SuperPoint) and PoSFeat \cite{li2022decoupling}. Besides, we take these recent matcher methods as baselines: DualRC-Net \cite{DualRC-Net}, SuperPoint+SuperGlue \cite{SuperGlue}, SuperPoint+SGMNet \cite{sgmnet} and Patch2Pix \cite{Patch2Pix}+HLOC\cite{HLOC}.

\begin{table}[h]
\centering
\renewcommand\arraystretch{1.5}
\caption{Evaluation results on Aachen Day-Night v1.1 dataset for visual localization.``High", ``Medium" and ``Coarse" respectively present $(0.25m, 2^{\circ})$, $(0.5m, 5^{\circ})$, and $(5m, 10^{\circ})$.}
\resizebox{.99\columnwidth}{!}
{
\begin{threeparttable}
\begin{tabular}{l|ccc}
    \hline
    \multirow{2}{*}{Methods} & \multicolumn{3}{c}{Aachen Day-Night v1.1} \\
    \cline{2-4}
            & High & Medium & Coarse \\
    \hline
    \hline
    R2D2-40K\tnote{1} \ \cite{R2D2}       & 71.2 & 86.9 & 97.9  \\
    ASLFeat-20K\tnote{2} \ \cite{aslfeat}    & 72.3 & 86.4 & 97.9 \\
    LISRD + SuperPoint + AdaLAM\tnote{1} \ \cite{LISRD} & 73.3 & 86.9 & 97.9 \\
    CNDesc + SuperPoint\tnote{3} \ \cite{cndesc} & 73.3 & 86.9 & 96.9 \\
    PoSFeat\tnote{1} \ \cite{li2022decoupling}            & \underline{73.8} & 87.4 & \textbf{98.4} \\
    
    \hline
    \hline
    DualRC-Net\tnote{1} \ \cite{DualRC-Net}             & 71.2 & 86.9 & 97.9 \\
    SuperPoint + SuperGlue\tnote{1} \ \cite{SuperGlue} & 73.3 & 88.0 & \textbf{98.4} \\
    SuperPoint + SGMNet\tnote{1} \ \cite{sgmnet}    & 72.3 & 85.3 & 97.9 \\
    Patch2Pix \cite{Patch2Pix} + HLOC\tnote{2} \ \cite{HLOC}                    & 72.3 & \underline{88.5} & 97.9 \\
    \hline
    \hline
    Ours  & \textbf{74.3} & \textbf{89.0} & \textbf{98.4} \\
    \hline
\end{tabular}
\begin{tablenotes}
\footnotesize
\item[1] Percentages reported at \cite{li2022decoupling}.
\item[2] Percentages reported at \href{https://www.visuallocalization.net/benchmark/}{https://www.visuallocalization.net} \cite{aachen}.
\item[3] Percentages reported at \cite{cndesc}.
\end{tablenotes}
\end{threeparttable}
}
\label{table2}
\end{table}

TABLE \ref{table2} exhibits two groups of numerical comparisons on Aachen Day-Night V1.1.
In comparisons, we could find that SCFeat often achieves a state-of-the-art (SOTA) performance in most cases in terms of ``High'', ``Medium'' and ``Coarse'', no matter if compared with extractor ones or matcher ones. For example, on ``Medium", the SOTA one of extractor methods, PoSFeat can get 87.4\%, and another SOTA one of matcher methods, Patch2Pix+HLOC can gain 88.5\%, while our SCFeat could reach to a new peak of 89.0\%.


\begin{table}[h]
\centering
\caption{Impact of Hyper-parameters: the evaluation comparisons on Aachen Day-Night v1.1 for visual localization, with maximal interest point number $20k$. ``High", ``Medium" and ``Coarse" present $(0.25m, 2^{\circ})$, $(0.5m, 5^{\circ})$, and $(5m, 10^{\circ})$, respectively. 
}
\renewcommand\arraystretch{1.2}
\resizebox{.99\columnwidth}{!}{
\begin{tabular}{lc|cc|ccc}
    \hline
    \multicolumn{2}{c|}{$N$} & \multicolumn{2}{c|}{$\epsilon_{s}$} & \multicolumn{3}{c}{Aachen Day-Night v1.1} \\
    \hline
    reference & query & reference & query & High & Medium & Coarse \\
    \hline
    1$\times$1 & 1$\times$1 & 1.0 & 1.0 & 73.3 & 88.0 & 98.4  \\
    
    3$\times$3 & 3$\times$3 & 0.3 & 0.3 & 73.3 & 86.4 & 98.4 \\
    
    3$\times$3 & 3$\times$3 & 0.5 & 0.5 & 73.3 & 88.5 & 98.4 \\
    
    5$\times$5 & 3$\times$3 & 0.5 & 0.5 & 72.3 & 88.5 & 98.4 \\
    
    5$\times$5 & 3$\times$3 & 1.0 & 0.5 & \textbf{74.3} & \textbf{89.0} & \textbf{98.4} \\
    
    \hline
\end{tabular}
}
\label{table11}
\end{table}

In addition, we conduct a group of extra comparison experiments to analyze the effects of the two main hyper-parameters, NMS size $N$ and score threshold $\epsilon_s$, on our SCFeat.
This group of comparison results are exhibited in TABLE \ref{table11}. It could be found that: i) our SCFeat is almost insensitive to these two parameters on ``Coarse'', and ii) SCFeat could achieve its optimal performance on both ``High'' and ``Medium'' when $N$ is $5\times5$ and $\epsilon_s=1.0$ for reference images, with $3\times3$ and $\epsilon_s=0.5$ for query ones. 

For example, the location precision on ``Coarse'' almost keeps a stable 98.4\% whatever values we set for both $N$ and $\epsilon_s$. Furthermore, when $N$ is a $5\times5$ with $\epsilon_s=1.0$ for reference images and $N$ is a $3\times3$ with $\epsilon_s=0.5$ for query ones, location precision could reach to 74.3\% and 89.0\% in terms of ``High'' and ``Medium'', respectively. 

\begin{figure}[h]
\centering
\includegraphics[width=0.99\columnwidth]{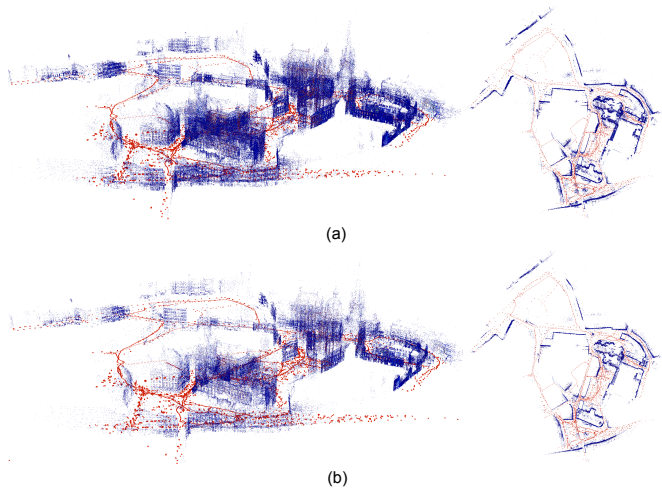}
\caption{
Impact of Hyper-parameters: the sparse 3D model of Aachen Day-Night v1.1 with different parameters: (a) a same NMS size of 3$\times$3 with score threshold 0.5 ($i.e.$, $\epsilon_{s}= 0.5$) for both reference images and query images, and (b) an NMS size of 5$\times$5 with $\epsilon_{s}= 1.0$) for reference images, and an NMS size of 3$\times$3 with $\epsilon_{s}= 0.5$ for query images.
}
\label{fig9}
\end{figure}

For NMS size $N$, its set depends on the resolution of the image. Due to the high resolution of the images in this dataset, the first group of parameters with a small NMS size ($i.e.$, $N$=1$\times$1) achieves a medium performance. 
Moreover, compared with query images, reference images in this dataset have higher resolutions, so the NMS size should be larger. On this foundation, in the last two sets of parameters, we respectively set the NMS size $N$ to 5$\times$5 and 3$\times$3 on reference images and query images. 
Unfortunately, we cannot see desired results in the fourth group of parameters. It only achieves 72.3\% on ``High'', 1\% lower than that by the third group of parameters. 
Meanwhile, the last group of parameters brings the best performance by increasing the $\epsilon_s$ for reference images. 
It indicates that $\epsilon_s$ should increase synchronously with $N$, for better performance. A possible reason is that larger NMS size has larger receptive fields, however it could reduce the accuracy of visual localization at the same time. In addition, when $\epsilon_s=0.3$, it achieves the lowest performance (86.4\%) on ``Medium". Intuitively, it is necessary to filter out the feature points of low confidence by raising the threshold of scores.


For a more intuitive sense of parameter effects on our SCFeat, we additionally visualize the sparse 3D models on this dataset.
Fig. \ref{fig9} illustrates two special experimental examples for TABLE \ref{table11}. 
Those models are reconstructed using COLMAP \cite{schoenberger2016sfm,schoenberger2016mvs} with features extracted by our SCFeat, and it will be used to do night-time images localization. 
For example, it is easy to find that the smaller the NMS size, the denser the distribution of features.
As shown in Fig. \ref{fig9} (a), when $N$ is set to 3$\times$3 on reference images, we observe that more features are registered. However, dense features are not friendly in feature localization for high-resolution images. Two main reasons are involved, one of which is that it usually has a smaller receptive field, which reduces the repeatability of features. The other reason is that a large number of dense features also lead to higher computation. Therefore, its performance is lower than the other, that is, unmatched NMS size will reduce the accuracy of feature localization.


To sum up, for low resolution of the image, we can set the NMS size to 1$\times$1. For medium resolution of the image, such as query images in this dataset, the recommended N is 3$\times$3. Furthermore, NMS size should be set to 5$\times$5 or larger for higher resolution of the image. For $\epsilon_s$, to ensure the quality of keypoints, the points with low confidence ($i.e.$, $\epsilon_s < 0.5$) will be filtered out. And $\epsilon_s$ often increases synchronously with NMS size, for better performance.

\subsection{3D Reconstruction}\label{3d}

\begin{table}[h]
\centering
\renewcommand\arraystretch{1.5}
\caption{The comparisons of sparse 3D reconstruction results on ETH local feature benchmark \cite{ETH}. Fig. \ref{fig31} shows the sparse 3D reconstructions of Madrid Metropolis, Gendarmenmarkt and Tower of London for our method.}
{
\begin{threeparttable}
\begin{tabular}{l|l|cccc}
    \hline
    \multirow{1}{*}{Subset} & \multirow{1}{*}{Methods} & \multirow{1}{*}{nReg} & \multirow{1}{*}{nPts}  & mTLen & mRErr \\
    \hline
    \hline
    Madrid & RootSIFT\tnote{1} \ \cite{hcrt:21} & 500 & 116k & 6.32 & \textbf{0.60} \\
    Metropolis & SuperPoint\tnote{1} \ \cite{SuperPoint} & 438 & 29k & 9.03 & 1.02 \\
    (1344 images) & D2-Net\tnote{1} \ \cite{D2net} & 501 & 84k & 6.33 & 1.28 \\
                  & ASLFeat\tnote{1} \ \cite{aslfeat} & 613 & 96k & 8.76 & 0.90 \\
                  & CAPS\tnote{1} \ \cite{caps} & \textbf{851} & 242k & 6.16 & 1.03 \\
                  & CoAM\tnote{1} \ \cite{CoAM} & 702 & 256k & 6.09 & 1.30 \\
                  & PoSFeat\tnote{1} \ \cite{li2022decoupling} & 419 & 72k & \underline{9.18} & 0.86 \\
                  \cdashline{2-6}
                  & Ours & 399 & 30k & \textbf{10.02} & \underline{0.84} \\
    \hline
    \hline
    Gendar- & RootSIFT\tnote{1} \ \cite{hcrt:21} & 1035 & 339k & 5.52 & \textbf{0.70} \\
    menmarkt & SuperPoint\tnote{1} \ \cite{SuperPoint} & 967 & 93k & 7.22 & 1.03 \\
    (1463 images) & D2-Net\tnote{1} \ \cite{D2net} & 1053 & 250k & 5.08 & 1.19 \\
                  & ASLFeat\tnote{1} \ \cite{aslfeat} & 1040 & 221k & \underline{8.72} & 1.00 \\
                  & CAPS\tnote{1} \ \cite{caps} & \textbf{1179} & 627k & 5.31 & 1.00 \\
                  & CoAM\tnote{1} \ \cite{CoAM} & 1072 & 570k & 6.60 & 1.34 \\
                  & PoSFeat\tnote{1} \ \cite{li2022decoupling} & 956 & 240k & 8.40 & \underline{0.92} \\
                  \cdashline{2-6}
                  & Ours & 917 & 108k & \textbf{9.78} & \underline{0.94} \\
    \hline
    \hline
     Tower of      & RootSIFT\tnote{1} \ \cite{hcrt:21} & 806 & 239k & 7.76 & \textbf{0.61} \\
     London        & SuperPoint\tnote{1} \ \cite{SuperPoint} & 681 & 52k & 8.67 & 0.96 \\
     (1576 images) & D2-Net\tnote{1} \ \cite{D2net} & 785 & 180k & 5.32 & 1.24 \\
                  & ASLFeat\tnote{1} \ \cite{aslfeat} & 821 & 222k & \textbf{12.52} & 0.92 \\
                  & CAPS\tnote{1} \ \cite{caps} & \textbf{1104} & 452k & 5.81 & 0.98 \\
                  & CoAM\tnote{1} \ \cite{CoAM} & 804 & 239k & 5.82 & 1.32 \\
                  & PoSFeat\tnote{1} \ \cite{li2022decoupling} & 778 & 262k & \underline{11.64} & 0.90 \\
                  \cdashline{2-6}
                  & Ours & 657 & 108k & \underline{11.62} & \underline{0.79} \\
    \hline
\end{tabular}
\begin{tablenotes}
\footnotesize
\item[1] Results reported at \cite{li2022decoupling}.
\end{tablenotes}
\end{threeparttable}
}
\label{table3}
\end{table}

Besides local feature matching and visual location, to verify the wide adaptability of our SCFeat, we continue to carry out a series of 3D reconstruction experiments on the ETH local features benchmark \cite{ETH}. It takes custom local features as input, and then relies on the COLMAP \cite{schoenberger2016sfm,schoenberger2016mvs} for 3D reconstruction.
In experiments, we use three medium-scale sub-datasets with a large number of different cameras and conditions, including {Madrid Metropolis}, {Gendarmenmarkt}, and {Tower of London}.

To evaluate the effects of different methods in 3D reconstruction task, we take the number of registered images (nReg), sparse 3D points (nPts), mean track lengths (mTLen) and mean re-projection error (mRErr) as four main performance metrics. In details, NMS size $N$ is set to 3$\times$3 and score map threshold $\epsilon_{s}=1.0$. In matching point pairs, we additionally apply a ratio test \cite{hcrt:21} with $\epsilon_{r}=0.80$ to achieve robust reconstruction. Moreover, the maximum number of keypoints is also limited to $20k$.
In experiments, we compare our SCFeat scheme quantitatively with seven recent baseline ones, including the RootSIFT \cite{hcrt:21}, SuperPoint \cite{SuperPoint}, D2-Net \cite{D2net}, ASLFeat \cite{aslfeat}, CAPS \cite{caps}, CoAM \cite{CoAM} and PosFeat \cite{li2022decoupling}.

TABLE \ref{table3} exhibits three groups of experimental comparison results on ETH.
By comparisons, it is easy to see that our SCFeat is overall comparable to these chosen baselines in supporting 3D reconstruction. In some cases, 
SCFeat could even obviously outperform others.

For example, on Madrid Metropolis our mTLen is 10.02 (the best one) and mRErr is 0.84 (the best one among all learning-based methods). It assists COLMAP to implement 3D reconstruction only using 
$30k$ spare points, generating a mRErr of 0.84. In contrast, as a SOTA one of local feature extraction, PoSFeat needs 
$72k$ points to support COLMAP for 3D reconstruction. 
It demonstrates that our keypoints could be more robust than those of PoSFeat, and thus easy to track across a large number of images. 

\begin{table}[h]
\centering
\caption{Impact of Hyper-parameters: The comparisons of sparse 3D reconstruction results on ETH local feature benchmark. The maximum interest point number is limited to $20k$. 
}
\renewcommand\arraystretch{1.2}
\resizebox{.99\columnwidth}{!}{
\begin{tabular}{l|ccc|cccc}
    \hline
    \multirow{1}{*}{Subset} & \multirow{1}{*}{$\epsilon_{r}$} & \multirow{1}{*}{$N$} & \multirow{1}{*}{$\epsilon_{s}$} & \multirow{1}{*}{nReg} & \multirow{1}{*}{nPts}  & mTLen & mRErr  \\
    \hline
    & 0.75 & 1$\times$1 & 1.0 & 158 & 12k & 6.34 & \textbf{0.49}\\
    Madrid & 0.80 & 3$\times$3 & 1.0 & 399 & 30k & \textbf{10.02} & \underline{0.84} \\
    Metropolis& 0.90 & 3$\times$3 & 0.5 & \underline{688} & \textbf{108k} & 9.35 & 1.05 \\
    & 0.92 & 3$\times$3 & 1.0 & \textbf{696} & \underline{89k} & \underline{9.42} & 1.05 \\
    
    \hline
    & 0.75 & 1$\times$1 & 1.0 & 605 & 51k & 6.95 & \textbf{0.66}\\
    Gendar-& 0.80 & 3$\times$3 & 1.0 & 917 & 108k & 9.78 & \underline{0.94} \\
    menmarkt& 0.90 & 3$\times$3 & 0.5 & \underline{1059} & \textbf{250k} & \underline{9.94} & 1.13 \\
    & 0.92 & 3$\times$3 & 1.0 & \textbf{1066} & \underline{192k} & \textbf{10.01} & 1.14 \\
    
    \hline
    & 0.75 & 1$\times$1 & 1.0 & 463 & 79k & 8.72 & \textbf{0.56}\\
    Tower of& 0.80 & 3$\times$3 & 1.0 & 657 & 108k & \textbf{11.62} & \underline{0.79} \\
    London& 0.90 & 3$\times$3 & 0.5 & \underline{878} & \textbf{233k} & \underline{11.39} & 1.03 \\
    & 0.92 & 3$\times$3 & 1.0 & \textbf{903} & \underline{185k} & 10.84 & 1.05 \\
    
    \hline
\end{tabular}
}
\label{table21}
\end{table}

\begin{figure}[!t]
\centering
\includegraphics[width=0.49\textwidth]{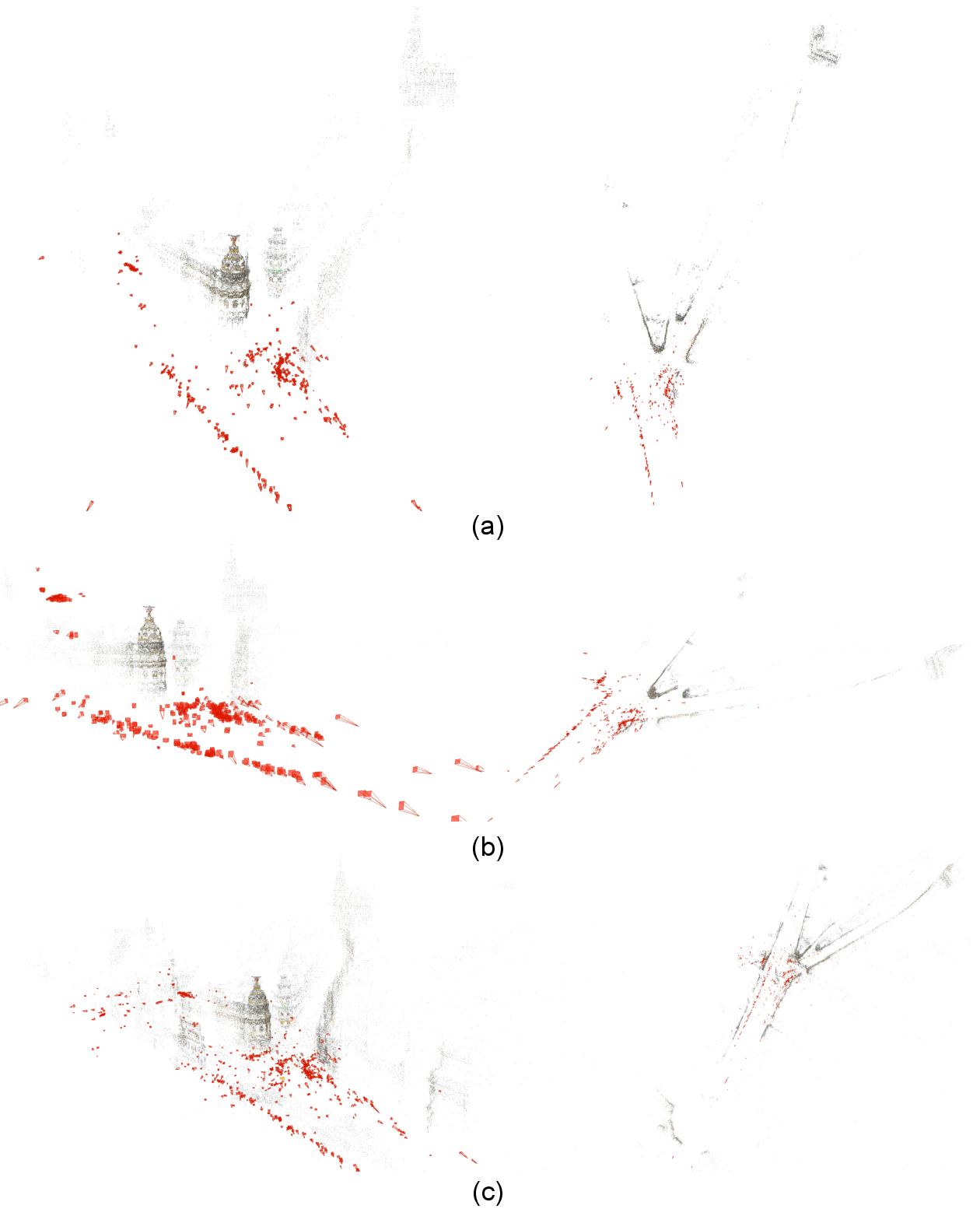}
\caption{Sparse 3D reconstructions of Madrid Metropolis for our SCFeat at different parameters. (a) NMS size $N$ is set to 1$\times$1, $\epsilon_{r}=0.75$, , $\epsilon_{s}=1.0$. (b) NMS size $N$ is set to 3$\times$3, $\epsilon_{r}=0.80$, $\epsilon_{s}=1.0$. (c) NMS size $N$ is set to 3$\times$3, $\epsilon_{r}=0.92$, $\epsilon_{s}=1.0$.}
\label{fig31}
\end{figure}

\begin{figure}[!t]
\centering
\includegraphics[width=0.49\textwidth]{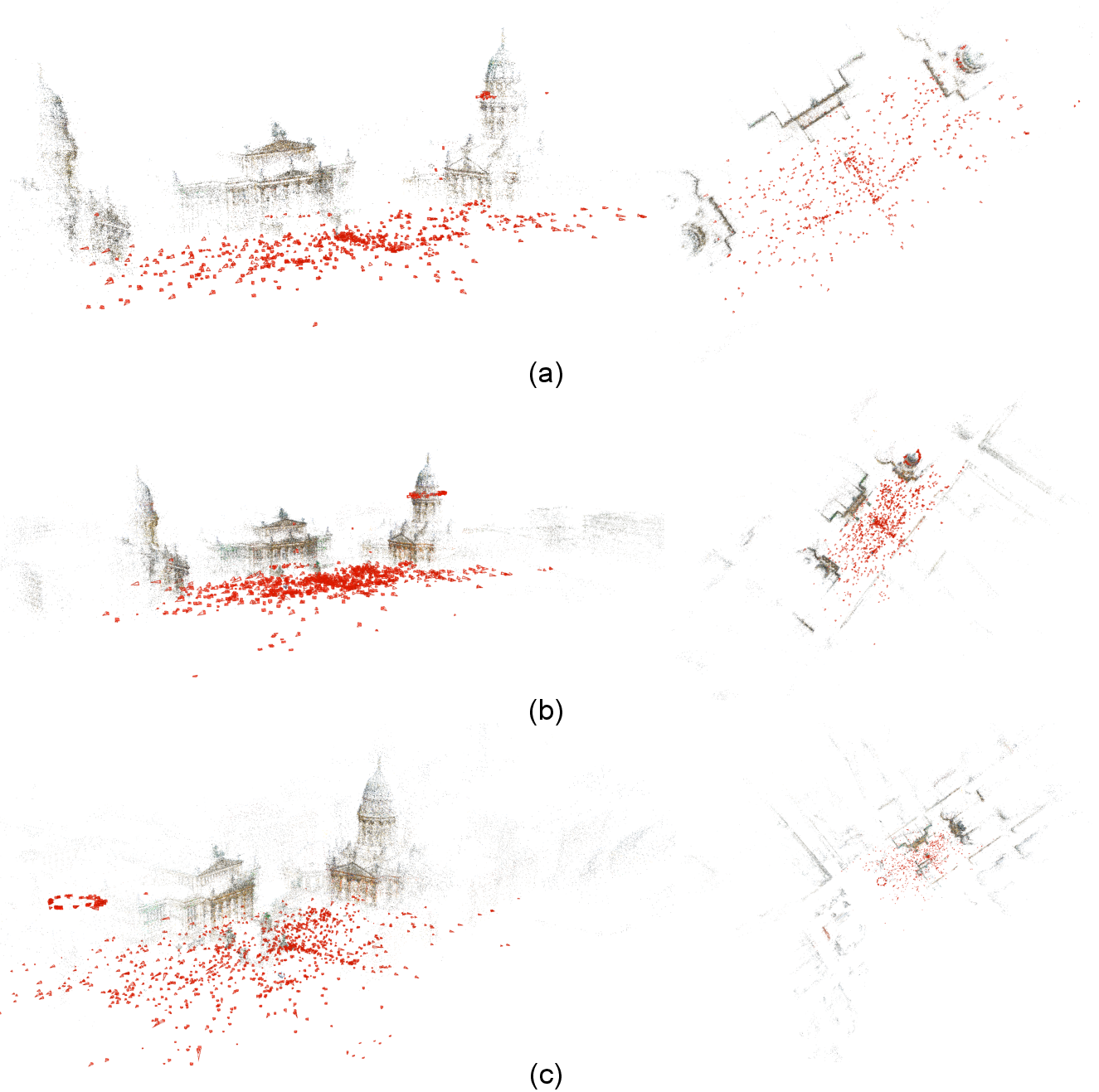}
\caption{Sparse 3D reconstructions on Gendarmenmarkt for our SCFeat at different parameters: (a) NMS size $N$ is set to 1$\times$1, $\epsilon_{r}=0.75$, $\epsilon_{s}=1.0$. (b) NMS size $N$ is set to 3$\times$3, $\epsilon_{r}=0.80$, $\epsilon_{s}=1.0$. (c) NMS size $N$ is set to 3$\times$3, $\epsilon_{r}=0.92$, $\epsilon_{s}=1.0$.}
\label{fig32}
\end{figure}

\begin{figure}[!t]
\centering
\includegraphics[width=0.49\textwidth]{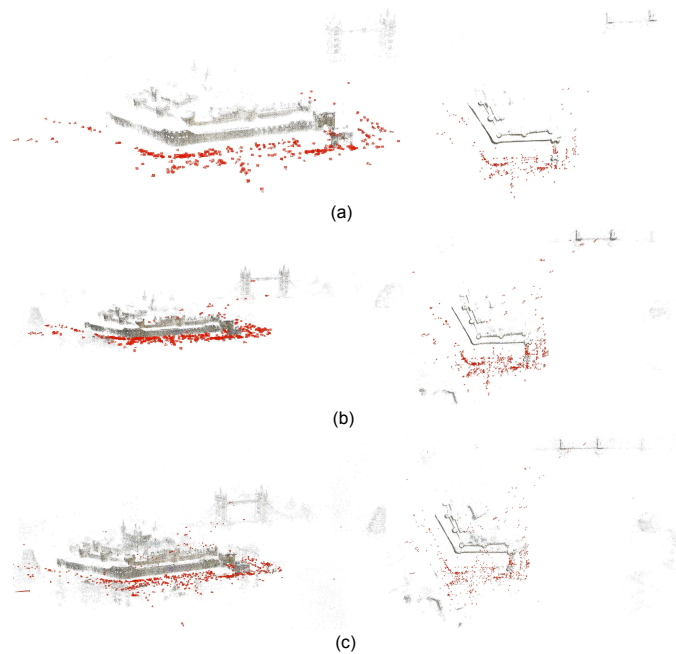}
\caption{Sparse 3D reconstructions of Tower of London for our SCFeat at different parameters. (a) NMS size $N$ is set to 1$\times$1, $\epsilon_{r}=0.75$, , $\epsilon_{s}=1.0$. (b) NMS size $N$ is set to 3$\times$3, $\epsilon_{r}=0.80$, $\epsilon_{s}=1.0$. (c) NMS size $N$ is set to 3$\times$3, $\epsilon_{r}=0.92$, $\epsilon_{s}=1.0$.}
\label{fig33}
\end{figure}

In TABLE \ref{table21}, we exhibit three groups of extra experiments, one group for one sub-dataset, to analyze the effects of NMS size $N$, score threshold $\epsilon_s$ and ratio test threshold $\epsilon_{r}$ on SCFeat performance. 
In the meanwhile, we additionally visualize one representative example of sparse 3D reconstruction for one sub-dataset. These three visualizations are exhibited in Fig. \ref{fig31}, Fig. \ref{fig32} and Fig. \ref{fig33}, respectively. 

From TABLE \ref{table21} and these three tables, we find that the number of nReg and nPts will gradually increase, while the mRErr will simultaneously decrease with the increase of $\epsilon_{r}$. In the meanwhile, it is no longer possible to meticulously distinguish the similar buildings on both the left and right sides of Gendarmenmarkt sample when $\epsilon_s=0.92$, as shown in Fig. \ref{fig32} (b) and (c).
Analyzed totally, it possibly means that a large threshold $\epsilon_{r}$ in our SCFeat could result in a large number of wrong matches. This phenomenon exactly re-confirms the conclusions in publication \cite{hcrt:21}. Therefore, as suggested in this publication, we set an appropriate threshold of $\epsilon_{r}=0.80$) in our optimal scheme of SCFeat parameters.

According to the findings observed in Sec. \ref{vl}, we thus set a suitable parameter of $\epsilon_s \geq 0.5$. For the size $N$ for NMS, because of the resolutions of the sample images in ETH, our SCFeat with the first group of parameters, $i.e.$, NMS size 1$\times$1, achieves a less nReg performance as speculated (only 158 registered images on Madrid Metropolis), although it achieves a better reconstruction mRErr. In the last three groups of parameters, due to a bigger NMS size of 3$\times$3, it achieves promising results as expected, which is quite consistent with the experiential inference as in Sec. \ref{vl}.

\section{Conclusions}
This paper proposes an augmented architecture for the weakly-supervised learning of local feature extraction. It proposes F2R-Backbone for local descriptors extraction. For more efficient local feature learning, it designs a coupling-bridge scheme of cross normalization between description network and detection network.
Moreover, an augmented detection network with peakiness measurement is devised for local feature detection, and a reward factor of fundamental matrix error is devised to optimize feature detection network training. Experiments shows that our scheme could usually obtain a state-of-the-art performance on image matching and visual localization. For 3D reconstruction, our scheme could also obtain competitive results. Moreover, ablation study further verifies that our F2R-Backbone, shared coupling-bridge strategy, and reward factor are effective in promoting the performance of feature extraction. Further work could be continued in terms of how to design faster and more efficient \emph{describe-then-detect} architecture. Besides, it is still worth investigating whether our SCFeat could better tackle some 3D vision tasks such as SfM and SLAM.

\bibliographystyle{unsrt}
\bibliography{mybibfile}

\begin{thebibliography}{10}

\bibitem{9261135}
Xin Yang, Zikang Yuan, Dongfu Zhu, Cheng Chi, Kun Li, and Chunyuan Liao.
\newblock Robust and efficient rgb-d slam in dynamic environments.
\newblock {\em IEEE Transactions on Multimedia}, 23:4208--4219, 2021.

\bibitem{9531062}
Bohong Yang, Wu~Ran, Lin Wang, Hong Lu, and Yi-Ping~Phoebe Chen.
\newblock Multi-classes and motion properties for concurrent visual slam in
  dynamic environments.
\newblock {\em IEEE Transactions on Multimedia}, 24:3947--3960, 2022.

\bibitem{9173732}
Xiaoxi Gong, Yuanpeng Liu, Qiaoyun Wu, Jiayi Huang, Hua Zong, and Jun Wang.
\newblock An accurate, robust visual odometry and detail-preserving
  reconstruction system.
\newblock {\em IEEE Transactions on Multimedia}, 23:2820--2832, 2021.

\bibitem{9527141}
Haoyuan Zhang, Lap-Pui Chau, and Danwei Wang.
\newblock Soft warping based unsupervised domain adaptation for stereo
  matching.
\newblock {\em IEEE Transactions on Multimedia}, 24:3835--3846, 2022.

\bibitem{9447920}
Yong Deng, Jimin Xiao, and Steven~Zhiying Zhou.
\newblock Tof and stereo data fusion using dynamic search range stereo
  matching.
\newblock {\em IEEE Transactions on Multimedia}, 24:2739--2751, 2022.

\bibitem{9817616}
Junna Gao, Dehui Kong, Shaofan Wang, Jinghua Li, and Baocai Yin.
\newblock Dasi: Learning domain adaptive shape impression for 3d object
  reconstruction.
\newblock {\em IEEE Transactions on Multimedia}, pages 1--15, 2022.

\bibitem{6359953}
Dimitrios~S. Alexiadis, Dimitrios Zarpalas, and Petros Daras.
\newblock Real-time, full 3-d reconstruction of moving foreground objects from
  multiple consumer depth cameras.
\newblock {\em IEEE Transactions on Multimedia}, 15(2):339--358, 2013.

\bibitem{8962030}
Chenggang Yan, Biyao Shao, Hao Zhao, Ruixin Ning, Yongdong Zhang, and Feng Xu.
\newblock 3d room layout estimation from a single rgb image.
\newblock {\em IEEE Transactions on Multimedia}, 22(11):3014--3024, 2020.

\bibitem{D2net}
Mihai Dusmanu, Ignacio Rocco, Tomas Pajdla, Marc Pollefeys, Josef Sivic,
  Akihiko Torii, and Torsten Sattler.
\newblock D2-net: A trainable cnn for joint description and detection of local
  features.
\newblock In {\em 2019 IEEE/CVF Conference on Computer Vision and Pattern
  Recognition (CVPR)}, pages 8084--8093, 2019.

\bibitem{hcrt:21}
David~G. Lowe.
\newblock Distinctive image features from scale-invariant keypoints.
\newblock {\em International Journal of Computer Vision}, 60:1573--1405, 2004.

\bibitem{BRISK}
Stefan Leutenegger, Margarita Chli, and Roland~Y. Siegwart.
\newblock Brisk: Binary robust invariant scalable keypoints.
\newblock In {\em 2011 International Conference on Computer Vision}, pages
  2548--2555, 2011.

\bibitem{Harris1988ACC}
Christopher~G. Harris and M.~J. Stephens.
\newblock A combined corner and edge detector.
\newblock In {\em Alvey Vision Conference}, 1988.

\bibitem{hcrt:28}
J~Matas, O~Chum, M~Urban, and T~Pajdla.
\newblock Robust wide-baseline stereo from maximally stable extremal regions.
\newblock {\em Image and Vision Computing}, 22(10):761--767, 2004.
\newblock British Machine Vision Computing 2002.

\bibitem{SURF}
Herbert Bay, Tinne Tuytelaars, and Luc Van~Gool.
\newblock Surf: Speeded up robust features.
\newblock In Ale{\v{s}} Leonardis, Horst Bischof, and Axel Pinz, editors, {\em
  Computer Vision -- ECCV 2006}, pages 404--417, Berlin, Heidelberg, 2006.
  Springer Berlin Heidelberg.

\bibitem{ORB}
Ethan Rublee, Vincent Rabaud, Kurt Konolige, and Gary Bradski.
\newblock Orb: An efficient alternative to sift or surf.
\newblock In {\em 2011 International Conference on Computer Vision}, pages
  2564--2571, 2011.

\bibitem{BRIEF}
Michael Calonder, Vincent Lepetit, Mustafa Ozuysal, Tomasz Trzcinski, Christoph
  Strecha, and Pascal Fua.
\newblock Brief: Computing a local binary descriptor very fast.
\newblock {\em IEEE Transactions on Pattern Analysis and Machine Intelligence},
  34(7):1281--1298, 2012.

\bibitem{R2D2}
Jerome Revaud, Cesar De~Souza, Martin Humenberger, and Philippe Weinzaepfel.
\newblock R2d2: Reliable and repeatable detector and descriptor.
\newblock In {\em Advances in Neural Information Processing Systems},
  volume~32. Curran Associates, Inc., 2019.

\bibitem{SuperPoint}
Daniel DeTone, Tomasz Malisiewicz, and Andrew Rabinovich.
\newblock Superpoint: Self-supervised interest point detection and description.
\newblock In {\em 2018 IEEE/CVF Conference on Computer Vision and Pattern
  Recognition Workshops (CVPRW)}, pages 337--33712, 2018.

\bibitem{cndesc}
Changwei Wang, Rongtao Xu, Shibiao Xu, Weiliang Meng, and Xiaopeng Zhang.
\newblock Cndesc: Cross normalization for local descriptors learning.
\newblock {\em IEEE Transactions on Multimedia}, pages 1--1, 2022.

\bibitem{LIFT}
Kwang~Moo Yi, Eduard Trulls, Vincent Lepetit, and Pascal Fua.
\newblock Lift: Learned invariant feature transform.
\newblock In Bastian Leibe, Jiri Matas, Nicu Sebe, and Max Welling, editors,
  {\em Computer Vision -- ECCV 2016}, pages 467--483, Cham, 2016. Springer
  International Publishing.

\bibitem{LF-Net}
Yuki Ono, Eduard Trulls, Pascal Fua, and Kwang~Moo Yi.
\newblock Lf-net: Learning local features from images.
\newblock In {\em Proceedings of the 32nd International Conference on Neural
  Information Processing Systems}, NIPS'18, page 6237–6247, Red Hook, NY,
  USA, 2018. Curran Associates Inc.

\bibitem{SOSNet}
Yurun Tian, Xin Yu, Bin Fan, Fuchao Wu, Huub Heijnen, and Vassileios Balntas.
\newblock Sosnet: Second order similarity regularization for local descriptor
  learning.
\newblock In {\em 2019 IEEE/CVF Conference on Computer Vision and Pattern
  Recognition (CVPR)}, pages 11008--11017, 2019.

\bibitem{ContextDesc}
Zixin Luo, Tianwei Shen, Lei Zhou, Jiahui Zhang, Yao Yao, Shiwei Li, Tian Fang,
  and Long Quan.
\newblock Contextdesc: Local descriptor augmentation with cross-modality
  context.
\newblock In {\em 2019 IEEE/CVF Conference on Computer Vision and Pattern
  Recognition (CVPR)}, pages 2522--2531, 2019.

\bibitem{caps}
Qianqian Wang, Xiaowei Zhou, Bharath Hariharan, and Noah Snavely.
\newblock Learning feature descriptors using camera pose supervision.
\newblock In Andrea Vedaldi, Horst Bischof, Thomas Brox, and Jan-Michael Frahm,
  editors, {\em Computer Vision -- ECCV 2020}, pages 757--774, Cham, 2020.
  Springer International Publishing.

\bibitem{aslfeat}
Zixin Luo, Lei Zhou, Xuyang Bai, Hongkai Chen, Jiahui Zhang, Yao Yao, Shiwei
  Li, Tian Fang, and Long Quan.
\newblock Aslfeat: Learning local features of accurate shape and localization.
\newblock {\em Computer Vision and Pattern Recognition (CVPR)}, 2020.

\bibitem{disk}
Micha{\l} Tyszkiewicz, Pascal Fua, and Eduard Trulls.
\newblock Disk: Learning local features with policy gradient.
\newblock {\em Advances in Neural Information Processing Systems}, 33, 2020.

\bibitem{li2022decoupling}
Kunhong Li, Longguang Wang, Li~Liu, Qing Ran, Kai Xu, and Yulan Guo.
\newblock Decoupling makes weakly supervised local feature better.
\newblock In {\em Proceedings of the IEEE/CVF Conference on Computer Vision and
  Pattern Recognition (CVPR)}, pages 15838--15848, June 2022.

\bibitem{HPatches}
Vassileios Balntas, Karel Lenc, Andrea Vedaldi, and Krystian Mikolajczyk.
\newblock Hpatches: A benchmark and evaluation of handcrafted and learned local
  descriptors.
\newblock In {\em 2017 IEEE Conference on Computer Vision and Pattern
  Recognition (CVPR)}, pages 3852--3861, 2017.

\bibitem{aachen}
Scaramuzza~D. Zhang~Z, Sattler~T.
\newblock Reference pose generation for long-term visual localization via
  learned features and view synthesis.
\newblock {\em International journal of computer vision}, 129(4):821--844,
  2021.

\bibitem{ETH}
Johannes~L. Schönberger, Hans Hardmeier, Torsten Sattler, and Marc Pollefeys.
\newblock Comparative evaluation of hand-crafted and learned local features.
\newblock In {\em 2017 IEEE Conference on Computer Vision and Pattern
  Recognition (CVPR)}, pages 6959--6968, 2017.

\bibitem{1642666}
J.~Kannala and S.S. Brandt.
\newblock A generic camera model and calibration method for conventional,
  wide-angle, and fish-eye lenses.
\newblock {\em IEEE Transactions on Pattern Analysis and Machine Intelligence},
  28(8):1335--1340, 2006.

\bibitem{ORBSLAM3_TRO}
Carlos Campos, Richard Elvira, Juan~J. G\`omez, Jos\`e M.~M. Montiel, and
  Juan~D. Tard\`os.
\newblock {ORB-SLAM3}: An accurate open-source library for visual,
  visual-inertial and multi-map {SLAM}.
\newblock {\em IEEE Transactions on Robotics}, 37(6):1874--1890, 2021.

\bibitem{c:29}
Daniel~Ponsa Vassileios~Balntas, Edgar~Riba and Krystian Mikolajczyk.
\newblock Learning local feature descriptors with triplets and shallow
  convolutional neural networks.
\newblock In Edwin R.~Hancock Richard C.~Wilson and William A.~P. Smith,
  editors, {\em Proceedings of the British Machine Vision Conference (BMVC)},
  pages 119.1--119.11. BMVA Press, September 2016.

\bibitem{Quad-Networks}
Nikolay Savinov, Akihito Seki, L'Ubor Ladický, Torsten Sattler, and Marc
  Pollefeys.
\newblock Quad-networks: Unsupervised learning to rank for interest point
  detection.
\newblock In {\em 2017 IEEE Conference on Computer Vision and Pattern
  Recognition (CVPR)}, pages 3929--3937, 2017.

\bibitem{BN}
Sergey Ioffe and Christian Szegedy.
\newblock Batch normalization: Accelerating deep network training by reducing
  internal covariate shift.
\newblock In {\em Proceedings of the 32nd International Conference on
  International Conference on Machine Learning - Volume 37}, ICML'15, page
  448–456. JMLR.org, 2015.

\bibitem{GN}
Yuxin Wu and Kaiming He.
\newblock Group normalization.
\newblock In Vittorio Ferrari, Martial Hebert, Cristian Sminchisescu, and Yair
  Weiss, editors, {\em Computer Vision -- ECCV 2018}, pages 3--19, Cham, 2018.
  Springer International Publishing.

\bibitem{645754.668382}
Yann LeCun, L\'{e}on Bottou, Genevieve~B. Orr, and Klaus-Robert M\"{u}ller.
\newblock Efficient backprop.
\newblock In {\em Neural Networks: Tricks of the Trade, This Book is an
  Outgrowth of a 1996 NIPS Workshop}, page 9–50, Berlin, Heidelberg, 1998.
  Springer-Verlag.

\bibitem{LayerN}
Jimmy Ba, Jamie~Ryan Kiros, and Geoffrey~E. Hinton.
\newblock Layer normalization.
\newblock {\em ArXiv}, abs/1607.06450, 2016.

\bibitem{3305381.3305417}
David Balduzzi, Marcus Frean, Lennox Leary, J~P Lewis, Kurt Wan-Duo Ma, and
  Brian McWilliams.
\newblock The shattered gradients problem: If resnets are the answer, then what
  is the question?
\newblock In {\em Proceedings of the 34th International Conference on Machine
  Learning - Volume 70}, ICML'17, page 342–350. JMLR.org, 2017.

\bibitem{HyNet}
Yurun Tian, Axel Barroso~Laguna, Tony Ng, Vassileios Balntas, and Krystian
  Mikolajczyk.
\newblock Hynet: Learning local descriptor with hybrid similarity measure and
  triplet loss.
\newblock In H.~Larochelle, M.~Ranzato, R.~Hadsell, M.F. Balcan, and H.~Lin,
  editors, {\em Advances in Neural Information Processing Systems}, volume~33,
  pages 7401--7412. Curran Associates, Inc., 2020.

\bibitem{resnet}
Kaiming He, Xiangyu Zhang, Shaoqing Ren, and Jian Sun.
\newblock Deep residual learning for image recognition.
\newblock In {\em 2016 IEEE Conference on Computer Vision and Pattern
  Recognition (CVPR)}, pages 770--778, 2016.

\bibitem{IN}
Dmitry Ulyanov, Andrea Vedaldi, and Victor~S. Lempitsky.
\newblock Instance normalization: The missing ingredient for fast stylization.
\newblock {\em CoRR}, abs/1607.08022, 2016.

\bibitem{Hartley2004}
R.~I. Hartley and A.~Zisserman.
\newblock {\em Multiple View Geometry in Computer Vision}.
\newblock Cambridge University Press, ISBN: 0521540518, second edition, 2004.

\bibitem{NMS}
A.~Neubeck and L.~Van~Gool.
\newblock Efficient non-maximum suppression.
\newblock In {\em 18th International Conference on Pattern Recognition
  (ICPR'06)}, volume~3, pages 850--855, 2006.

\bibitem{MegaDepthLi18}
Zhengqi Li and Noah Snavely.
\newblock Megadepth: Learning single-view depth prediction from internet
  photos.
\newblock In {\em Computer Vision and Pattern Recognition (CVPR)}, 2018.

\bibitem{schoenberger2016sfm}
Johannes~Lutz Sch\"{o}nberger and Jan-Michael Frahm.
\newblock Structure-from-motion revisited.
\newblock In {\em Conference on Computer Vision and Pattern Recognition
  (CVPR)}, 2016.

\bibitem{schoenberger2016mvs}
Johannes~Lutz Sch\"{o}nberger, Enliang Zheng, Marc Pollefeys, and Jan-Michael
  Frahm.
\newblock Pixelwise view selection for unstructured multi-view stereo.
\newblock In {\em European Conference on Computer Vision (ECCV)}, 2016.

\bibitem{HesAffandSIFT}
Relja Arandjelović and Andrew Zisserman.
\newblock Three things everyone should know to improve object retrieval.
\newblock In {\em 2012 IEEE Conference on Computer Vision and Pattern
  Recognition}, pages 2911--2918, 2012.

\bibitem{HesAffNet}
Dmytro Mishkin, Filip Radenovi{\'{c}}, and Ji{\v{r}}i Matas.
\newblock Repeatability is not enough: Learning affine regions via
  discriminability.
\newblock In Vittorio Ferrari, Martial Hebert, Cristian Sminchisescu, and Yair
  Weiss, editors, {\em Computer Vision -- ECCV 2018}, pages 287--304, Cham,
  2018. Springer International Publishing.

\bibitem{HardNetpp}
Anastasiya Mishchuk, Dmytro Mishkin, Filip Radenović, and Jiri Matas.
\newblock Working hard to know your neighbor's margins:local descriptor
  learning loss.
\newblock {\em CoRR}, abs/1705.10872, 05 2017.

\bibitem{DELF}
Hyeonwoo Noh, Andre Araujo, Jack Sim, Tobias Weyand, and Bohyung Han.
\newblock Large-scale image retrieval with attentive deep local features.
\newblock In {\em 2017 IEEE International Conference on Computer Vision
  (ICCV)}, pages 3476--3485, 2017.

\bibitem{LISRD}
R{\'e}mi Pautrat, Viktor Larsson, Martin~R. Oswald, and Marc Pollefeys.
\newblock Online invariance selection for local feature descriptors.
\newblock In Andrea Vedaldi, Horst Bischof, Thomas Brox, and Jan-Michael Frahm,
  editors, {\em Computer Vision -- ECCV 2020}, pages 707--724, Cham, 2020.
  Springer International Publishing.

\bibitem{cavalli2020handcrafted}
Luca Cavalli, Viktor Larsson, Martin~Ralf Oswald, Torsten Sattler, and Marc
  Pollefeys.
\newblock Handcrafted outlier detection revisited.
\newblock In {\em European Conference on Computer Vision}, 2020.

\bibitem{DualRC-Net}
Xinghui Li, Kai Han, Shuda Li, and Victor Prisacariu.
\newblock Dual-resolution correspondence networks.
\newblock In {\em Proceedings of the 34th International Conference on Neural
  Information Processing Systems}, NIPS'20, Red Hook, NY, USA, 2020. Curran
  Associates Inc.

\bibitem{SuperGlue}
Paul-Edouard Sarlin, Daniel DeTone, Tomasz Malisiewicz, and Andrew Rabinovich.
\newblock Superglue: Learning feature matching with graph neural networks.
\newblock In {\em 2020 IEEE/CVF Conference on Computer Vision and Pattern
  Recognition (CVPR)}, pages 4937--4946, 2020.

\bibitem{sgmnet}
Hongkai Chen, Zixin Luo, Jiahui Zhang, Lei Zhou, Xuyang Bai, Zeyu Hu, Chiew-Lan
  Tai, and Long Quan.
\newblock Learning to match features with seeded graph matching network.
\newblock {\em International Conference on Computer Vision (ICCV)}, 2021.

\bibitem{Patch2Pix}
Qunjie Zhou, Torsten Sattler, and Laura Leal-Taixé.
\newblock Patch2pix: Epipolar-guided pixel-level correspondences.
\newblock In {\em 2021 IEEE/CVF Conference on Computer Vision and Pattern
  Recognition (CVPR)}, pages 4667--4676, 2021.

\bibitem{HLOC}
Paul-Edouard Sarlin, Cesar Cadena, Roland Siegwart, and Marcin Dymczyk.
\newblock From coarse to fine: Robust hierarchical localization at large scale.
\newblock In {\em CVPR}, 2019.

\bibitem{CoAM}
Olivia Wiles, Sébastien Ehrhardt, and Andrew Zisserman.
\newblock Co-attention for conditioned image matching.
\newblock In {\em 2021 IEEE/CVF Conference on Computer Vision and Pattern
  Recognition (CVPR)}, pages 15915--15924, 2021.

\end{thebibliography}


 




\vfill

\end{document}